\documentclass[mnsc]{informs4}
\usepackage{eqndefns-left} 
\usepackage[titletoc,title]{appendix}
\RequirePackage{tgtermes}
\RequirePackage{newtxtext}
\RequirePackage{newtxmath}
\RequirePackage{bm}
\RequirePackage{endnotes}

\OneAndAHalfSpacedXII 

\usepackage[ruled]{algorithm2e} 
\usepackage{algpseudocode}
\usepackage{tikz}
\usetikzlibrary{decorations.pathreplacing}

\usepackage{natbib}
 \bibpunct[, ]{(}{)}{,}{a}{}{,}%
 \def\bibfont{\small}%
\EquationsNumberedThrough    

\TheoremsNumberedThrough     
\ECRepeatTheorems  %

\MANUSCRIPTNO{MNSC-0001-2024.00}

\usepackage{subcaption}

\newcommand{\E}{\mathbb{E}}

\usepackage{thmtools,thm-restate}  

\newenvironment{customproof}[1][Proof]{\par\noindent\textit{\textbf{#1.}} }{\hfill$\square$\par}

\usepackage{tcolorbox}
\usepackage{xcolor}
\begin{document}


\RUNAUTHOR{Ge, Bastani and Bastani}

\RUNTITLE{Rethinking Algorithmic Fairness for Human-AI Collaboration}

\TITLE{Rethinking Algorithmic Fairness for Human-AI Collaboration}

\ARTICLEAUTHORS{%
\AUTHOR{Haosen Ge}
\AFF{Wharton AI \& Analytics Initiative, 
Wharton School, University of Pennsylvania
\EMAIL{hge@wharton.upenn.edu}}

\AUTHOR{Hamsa Bastani}
\AFF{Department of Operations, Information, and Decisions,
Wharton School, University of Pennsylvania, \EMAIL{hamsab@wharton.upenn.edu}}

\AUTHOR{Osbert Bastani}
\AFF{Department of Computer and Information Science, University of Pennsylvania, \EMAIL{obastani@seas.upenn.edu}}
} 

\ABSTRACT{%
Existing approaches to algorithmic fairness aim to ensure equitable outcomes \emph{if} human decision-makers comply perfectly with algorithmic decisions. However, perfect compliance with the algorithm is rarely a reality or even a desirable outcome in human-AI collaboration. Yet, recent studies have shown that selective compliance with fair algorithms can \textit{amplify} discrimination relative to the prior human policy. As a consequence, ensuring equitable outcomes requires fundamentally different algorithmic design principles that ensure robustness to the decision-maker's (a priori unknown) compliance pattern. We define the notion of \textit{compliance-robustly} fair algorithmic recommendations that are guaranteed to (weakly) improve fairness in decisions, regardless of the human's compliance pattern. We propose a simple optimization strategy to identify the best performance-improving compliance-robustly fair policy.  However, we show that it may be infeasible to design algorithmic recommendations that are simultaneously fair in isolation, compliance-robustly fair, and more accurate than the human policy; thus, if our goal is to improve the equity and accuracy of human-AI collaboration, it may not be desirable to enforce traditional algorithmic fairness constraints. We illustrate the value of our approach on criminal sentencing data before and after the introduction of an algorithmic risk assessment tool in Virginia.
}%





\maketitle


\section{Introduction}
As machine learning algorithms are increasingly deployed in high-stakes settings (e.g., healthcare, finance, justice), it has become imperative to understand the fairness implications of algorithmic decision-making on protected groups. As a consequence, a wealth of work has sought to define algorithmic fairness~\citep{dwork2012fairness,hardt2016equality,corbett2018measure,kleinberg2016inherent,chen2023algorithmic} and learn machine learning-based policies that satisfy fairness constraints to ensure equitable outcomes across protected groups~\citep{kim2019multiaccuracy,bastani2022practical,kearns2019empirical,joseph2016fairness,basu2023use}.

Most of this work focuses on whether the algorithm makes fair decisions \textit{in isolation}. Yet, these algorithms are rarely used in high-stakes settings without human oversight, since there are still considerable legal and regulatory challenges to full automation. Moreover, many believe that human-AI collaboration is superior to full automation because human experts may have auxiliary information that can help correct the mistakes of algorithms, producing better decisions than the human or algorithm alone. For example, while many powerful AI systems have been developed for diagnosing medical images \citep{esteva2017dermatologist}, the Center for Medicare and Medicaid Services only allows AI systems to \textit{assist} medical experts with diagnosis \citep{rajpurkar2022ai}.

However, human-AI collaboration introduces new complexities---the overall outcomes now depend not only on the algorithmic recommendations, but also on the subset of individuals for whom the human decision-maker complies with the algorithmic recommendation. Recent case studies have shown mixed results on whether human-AI collaboration actually improves decision accuracy \citep{campero2022test,ahn2024impact} or fairness \citep{van2019algorithms}. For instance, a recent experiment examines diagnostic quality when radiologists are assisted by AI models \citep{agarwal2023combining}. The authors find that, although the AI models are substantially more accurate than radiologists, access to AI assistance does not improve diagnostic quality on average; the authors show that this is due to \textit{selective compliance} of the algorithmic recommendations by humans, which they hypothesize is driven by improper Bayesian updating. Similarly, a recent study evaluates the impact of algorithmic risk assessment on judges' sentencing decisions in Virginia courts \citep{stevenson2022algorithmic, van2019algorithms}. Although risk assessment promised fairer outcomes \citep{kleinberg2018human}, the authors find that it brought no detectable benefits in terms of public safety or reduced incarceration; in fact, racial disparities \textit{increased} in the subset of courts where risk assessment appears most influential. Once again, the mismatch is driven by selective compliance to algorithmic recommendations, which appears to be at least partly driven by conflicting objectives between judges and the algorithm (e.g., judges are more lenient towards younger defendants). Selective compliance has significant fairness implications, e.g., in this case, the authors note that ``judges were more likely to sentence leniently for white defendants with high-risk scores than for black defendants with the same score.''
These case studies make it clear that ensuring equitable outcomes in human-AI collaboration requires accounting for humans' complex and unexpected compliance patterns. 


To resolve this state of affairs, we introduce the notion of \textit{compliance-robust} algorithms---i.e., algorithmic decision policies that are guaranteed to (weakly) improve fairness in final outcomes, regardless of the human's (unknown) compliance pattern. In particular, given a human decision-maker and her policy (without access to AI assistance), we characterize the class of algorithmic recommendations that never result in collaborative final outcomes that are less fair than the pre-existing human policy, even if the decision-maker's compliance pattern is adversarial. Next, we prove that there exists considerable tension between traditional algorithmic fairness and compliance-robust fairness. Unless the true data-generating process is itself perfectly fair, it can be infeasible to design an algorithmic policy that is fair in isolation, compliance-robustly fair, and more accurate than the human-only policy, implying that compliance-robust fairness imposes fundamentally different constraints compared to traditional fairness. This raises the question of whether traditional fairness is even a desirable constraint to enforce for human-AI collaboration---if the goal is to improve fairness and accuracy in human-AI collaboration outcomes, it may be preferable to design an algorithmic policy that is accurate and compliance-robustly fair, but not necessarily fair in isolation. 

Lastly, we use Virginia court sentencing data---leveraging variation from the introduction of an algorithmic risk assessment tool in 2002, as proposed in \cite{stevenson2022algorithmic}---to simulate the performance and fairness of compliance-robust policies versus natural baseline policies. We find that a compliance-robust policy performs favorably, both in terms of performance and fairness, across all 170 judges who exhibit very different compliance behaviors.

\subsection{Related Literature}

There has been a long line of work studying fairness for algorithms in isolation of human decision-makers. Much of this work has focused on mathematical definitions of fairness~\citep{dwork2012fairness,hardt2016equality,kleinberg2016inherent,corbett2018measure} or on operationalizing these definitions in practice. For example, \citet{kallus2022assessing} study the practical challenge of (at least partially) ensuring fairness when the protected class membership is not observed, and \citet{cai2020fair} propose selectively acquiring costly additional information on individuals to improve fairness in downstream outcomes. Fairness is especially important for resource allocation problems. To this end, \citet{manshadi2023fair} derive a dynamic allocation policy that satisfies ex-ante and ex-post fairness constraints, and \citet{mulvany2021fair} propose going beyond the classic $c\mu$-rule to achieve fair scheduling of heterogeneous populations. However, in many practical applications, algorithmic predictions are shown to human decision-makers, who then make the final decision---this setting is the focus of our work.

There has been a great deal of recent interest in human-AI collaboration, but much of it focuses on improving performance rather than ensuring fairness. For instance, \citet{wang2024friend} examine how work experience influences the complementarity between humans and AI, finding that senior workers often benefit less from AI than junior workers, potentially because they trust AI less. \citet{tong2021janus} demonstrate a net positive effect on worker performance when AI is used to provide performance feedback, while \citet{bastani2021improving} investigate how interpretable ``tips" can optimally improve performance. \citet{balakrishnan2025human} find that human decision-makers display naïve advice-weighting behavior, harming outcomes. More closely related to our work, a growing literature examines how workers \textit{perceive} AI’s fairness. \citet{newman2020eliminating} find that workers view AI-based performance evaluations as less fair than those made by humans, largely because AI is seen as unable to incorporate sufficient contextual and qualitative information. By contrast, \citet{bai2022impacts} argues that workers may perceive AI as fairer than human decision-makers when an equality motive dominates---in a field experiment, warehouse workers perceived AI task assignments as more equitable. We build on this literature by examining AI's impact on actual fairness outcomes under selective compliance behaviors.

Our work is motivated by recent papers demonstrating a gap between fairness of an algorithm and fairness in human-AI collaboration. For example, using Kentucky court data,~\citet{albright2019if} show that the introduction of a risk assessment tool increased racial disparities in initial bond decisions because judges were more likely to override the tool’s recommendations for Black defendants than for otherwise similar White defendants. \citet{hoffman2018discretion} examine whether human discretion improves hiring outcomes when managers are assisted by job-testing technologies. They find that managers who appear to hire against test recommendations tend to make worse average hires, suggesting that managers are often biased or mistaken when exercising discretion. Subsequent work has sought to theoretically characterize fairness in human-AI collaboration. For instance, \citet{morgan2019paradoxes} show that a fair algorithmic policy cannot prevent humans from exploiting its recommendations in discriminatory ways. In fact, as we show, it is often impossible to design an algorithm that is both fair and robust to such exploitation. In contrast, our goal is to design algorithms that provide provable fairness guarantees for human-AI collaboration. \citet{gillis2021fairness} study why and how humans deviate from algorithmic recommendations using a Bayesian persuasion framework. They investigate potential remedies such as disclosing information about protected groups (along with the recommendation) or withholding recommendations altogether. These approaches require knowledge of how the human may react to the recommendation, which is often unknown a priori. Our compliance-robust fairness property guarantees fairness for arbitrary human compliance strategies.

More broadly, our work has a similar motivation to papers studying how to steer human decision-making to achieve some goal, although they focus on improvimg accuracy rather than fairness. For example, \citet{xu2023decision} use a game-theoretic framework to study when it is possible to use algorithms to ``control'' strategic human decision-makers to achieve the most desirable outcome; \citet{alur2024human} propose an algorithmic framework that provably improves AI's prediction accuracy by incorporating human expertise, even under imperfect human compliance.

\section{Problem Formulation}

\paragraph{Human-AI decision-making.} We first introduce some notation to formalize the human-AI decision-making problem, as well as our definitions of traditional fairness, compliance-robust fairness, and performance.

Consider a decision-making problem where each individual is associated with a type $x\in\mathcal{X}=[k]=\{1,...,k\}$ (e.g., education, prior defaults), a protected attribute $a \in \mathcal{A}=\{0,1\}$ (e.g., gender), and a true outcome $y \in \mathcal{Y}=\{0,1\}$ (e.g., whether they can repay a loan).\footnote{Note $\mathcal{X}$ can encompass multiple categorical features that are ``flattened'' into a single dimension. For mathematical simplicity, we restrict to categorical features (i.e., $\mathcal{X}$ has finite possible values), a binary protected attribute, and a binary outcome. Our results straightforwardly extend to protected attributes with multiple classes, but allowing for continuous features and outcomes requires modifying the primary fairness definition we use \cite{hardt2016equality} by introducing slack variables.} Let $\mathbb{P}(x,a,y)$ over $\mathcal{X}\times\mathcal{A}\times\mathcal{Y}$ denote the joint distribution of the types, protected attributes, and outcomes of individuals for whom decisions must be made (we will not require knowledge of $\mathbb{P}(x,a,y)$ to construct policies). We make the following assumption that our population has good ``coverage'' across variables/outcomes:
\begin{assumption}
\label{assump:probability}
We have $\mathbb{P}(x,a,y)>0$ for all $x\in\mathcal{X}$, $a\in\mathcal{A}$, and $y\in\mathcal{Y}$.
\end{assumption}

\begin{definition}
A \emph{decision-making policy} is a mapping $\pi:\mathcal{X}\times\mathcal{A}\to[0,1]$ that maps each feature-attribute pair to a probability $\pi(x,a)$; then, the decision is $\hat{y}\sim\text{Bernoulli}(\pi(x,a))$.
\end{definition}
We use the commonly employed Bernoulli distribution, but it suffices for $\mathbb{P}[\hat{y}=1]$ to simply be increasing in $\pi(x,a)$.

The algorithm designer's goal is for the decision to equal the true outcome---i.e., $\hat{y}=y$ (e.g., we would ideally give each individual a loan if and only if they will repay the loan).\footnote{Note that the human's objective may vary, e.g., judges are more lenient towards younger defendants \cite{van2019algorithms}, but algorithm designers are typically restricted to predicting outcomes observed in the training data.} We consider a human decision-maker represented by a policy $\pi_H$ (without access to algorithmic assistance). When given access to recommendations from an algorithmic policy $\pi_A$, the human instead makes decisions according to a \emph{compliance function} $c: \mathcal{X} \times \mathcal{A} \mapsto \{0,1\}$, where $c(x,a) = 1$ indicates that the human adopts the algorithmic decision for individuals $(x,a)$. Then, the joint human-AI policy will be
\begin{align*}
\pi_C(x,a)
&=\begin{cases}
\pi_A(x,a)&\text{if }c(x,a)=1 \\
\pi_H(x,a)&\text{otherwise}.
\end{cases}
\end{align*}
Note that $\pi_H$ can be estimated via supervised learning on historical decision-making data in the absence of algorithmic recommendations; in contrast, the compliance function $c$ cannot be learned until an algorithm $\pi_A$ has already been deployed, potentially with poor consequences. Thus, we assume knowledge of $\pi_H$, but study compliance-robust policies that do not require any knowledge of $c$. We show that our results extend straightforwardly to (i) compliance functions that depend on the algorithmic recommendation itself (e.g., a human is more likely to comply when the algorithmic recommendation $\pi_A(x,a)$ is similar to their own judgment $\pi_H(x,a)$), (ii) stochastic (rather than deterministic) compliance functions, or (iii) partial compliance (e.g., the decision-maker takes a weighted average of their own judgment and the AI recommendation). We discuss these alternative specifications after Theorem~\ref{thm:main}.

\paragraph{Fairness.}
We primarily analyze the well-studied notion of ``equality of opportunity'' \citep{hardt2016equality, kleinberg2016inherent}, which requires that, for any chosen decision policy $\pi$, the true positive rates for each protected group should be equal:
\[\mathbb{P}[\hat{y}=1 \mid y=1, a=0] = \mathbb{P}[\hat{y}=1 \mid y=1, a=1].\]
In other words, on average, deserving individuals ($y=1$) should have the same likelihood of access to the intervention ($\hat{y}=1$) regardless of their protected group status ($a\in\{0,1\}$). In Appendix \ref{general_fairness}, we show that the qualitative challenges that we illustrate in this paper arise for a very general class of fairness definitions, subsuming demographic parity \citep{calders2009building,zliobaite2015relation} and equalized odds \citep{hardt2016equality,chen2023algorithmic}.

For a policy $\pi$, we then marginalize out the types $x$ to obtain the average score for subgroup $a$ as
\begin{align*}
\overline{\pi}(a) = \sum_{x\in \mathcal{X}}\pi(x,a)\mathbb{P}(x\mid a,y=1).
\end{align*}
Traditional algorithmic fairness would require the algorithmic policy $\pi_A$ to satisfy $\overline{\pi}_A(0) = \overline{\pi}_A(1)$, without accounting for the human policy $\pi_H$ or the compliance function $c$.

Next, without loss of generality, we assume Group 1 is better off than Group 0 in terms of ``opportunity'' under the human-alone policy:
\begin{assumption}
\label{assump:group}
We have $\overline{\pi}_H(1)\ge\overline{\pi}_H(0)$.
\end{assumption}
We now introduce some definitions. Let $\alpha$ be the slack in group fairness for a policy $\pi$:
\begin{align*}
\alpha(\pi) = |\overline{\pi}(1) - \overline{\pi}(0)|.
\end{align*}

\begin{definition}
\rm
We say an algorithmic policy $\pi_A$ \emph{reduces fairness} under compliance function $c$ if the resulting human-AI policy $\pi_C$ satisfies $\alpha(\pi_C)>\alpha(\pi_H)$.
\end{definition}

Note that a human decision-maker can always choose to ignore all algorithmic advice (i.e., $c(x,a) = 0$ for all $x$ and $a$) resulting in the human's policy ($\pi_C = \pi_H$)---then, if $\pi_H$ is unfair, no choice of $\pi_A$ can guarantee a fair $\pi_C$. Thus, when designing $\pi_A$, we can at most demand that we do not reduce unfairness relative to the existing human policy $\pi_H$.

\begin{definition}
\rm
Given $\pi_H$, an algorithmic policy $\pi_A$ is \emph{compliance-robustly fair} if there does not exist \textit{any} compliance function $c$ that reduces fairness for $\pi_A$.
\end{definition}

Let $\Pi_{\text{fair}}$ be the set of compliance-robustly fair policies; note that these policies need not be fair in the traditional algorithmic fairness sense. We will characterize $\Pi_{\text{fair}}$ in the next section.

\paragraph{Performance.} Algorithmic assistance often aims to not only improve fairness but also the \textit{accuracy} of decisions. Ideally, we would produce compliance-robustly fair recommendations that improve performance relative to the human policy. To define performance, we consider a loss function $\ell:[0,1]\times\mathcal{Y}\to\mathbb{R}$, and define the expected loss 
\begin{align*}
L(\pi)=\E[\ell(\pi(x,a),y)].
\end{align*}
Let the performance-maximizing (but possibly unfair) optimal policy be
\begin{align*}
\pi_* = \argmin_\pi L(\pi),
\end{align*}
and the highest performing compliance-robustly fair policy be
\begin{align*}
\pi_0=\operatorname*{\arg\min}_{\pi\in\Pi_{\text{fair}}}L(\pi).
\end{align*}

For analysis, we impose the following mild assumption on our loss:
\begin{definition}
\rm
We say a policy $\pi'$ has \emph{higher deviation} than a second policy $\pi$ if for all $x\in\mathcal{X},a\in\mathcal{A}$, if $\pi(x,a)\ge\pi_*(x,a)$, then $\pi'(x,a)\ge\pi(x,a)$, and if $\pi(x,a)\le\pi_*(x,a)$, then $\pi'(x,a)\le\pi(x,a)$. We say the deviation is \emph{strictly higher} if the inequality is strict for any $x\in\mathcal{X},a\in\mathcal{A}$.
\end{definition}
\begin{assumption}
\label{assump:loss}
For any policies $\pi,\pi'$, if $\pi'$ has higher deviation than $\pi$, then $L(\pi')\ge L(\pi)$; furthermore, if the deviation is strictly higher, then $L(\pi')>L(\pi)$.
\end{assumption}
In other words, if $\pi'$ always deviates farther from $\pi_*$ than $\pi$ (i.e., for every $x$ and $a$), then $\pi'$ has higher expected loss. It can be easily checked that common loss functions (e.g., mean squared error, mean absolute error, cross entropy) satisfy the above definition. It is worth noting that we do not assume the loss is symmetric---i.e., if $\pi$ and $\pi'$ are on different sides of $\pi_*$ for any $x,a$ pair, this assumption does not say anything about which one attains a lower loss. 

\section{Characterization of Compliance-Robust Fairness}
\label{section:main}
\begin{figure}[t]
\centering
\begin{subfigure}{0.5\textwidth}
\centering
\begin{tikzpicture}[scale=1.5,font=\fontsize{8}{8}\selectfont]
\draw[->] (0,0) -- (3,0) node[right] {$a$};
\node[] at (0.7,-0.2) {$a=0$};
\node[] at (2.2,-0.2) {$a=1$};
\draw[->] (0,0) -- (0,3) node[above] {$\overline{\pi}(a)$};
\draw[dotted] (1.5,0) -- (1.5,3);
\draw[-, red] (0,0.8) -- (1.5,0.8);
\draw[-, red] (1.5,1.5) -- (3,1.5) node[xshift=0.3cm] {$\pi_H$};
\draw[dashed, blue] (0,2.5) -- (3,2.5) node[xshift=0.3cm] {$\pi_A$};
\draw [decorate,decoration={brace,amplitude=8pt},xshift=-0.01cm,yshift=0pt]
(1.5,0.8) -- (1.5,1.5) node [black,midway,xshift=-0.8cm]
{\footnotesize $\alpha(\pi_H)$};
\draw [-stealth](2.25,1.5) -- (2.25,2.5);
\end{tikzpicture}
\caption{Fairness reduced if $c(x,0)=0$ and $c(x,1)=1$}
\label{fig:example:reduce_fairness}
\end{subfigure}%
\begin{subfigure}{0.5\textwidth}
\centering
\begin{tikzpicture}[scale=1.5,font=\fontsize{8}{8}\selectfont]
\draw[->] (0,0) -- (3,0) node[right] {$a$};
\node[] at (0.7,-0.2) {$a=0$};
\node[] at (2.2,-0.2) {$a=1$};
\draw[->] (0,0) -- (0,3) node[above] {$\overline{\pi}(a)$};
\draw[dotted] (1.5,0) -- (1.5,3);
\draw[-, red] (0,0.8) -- (1.5,0.8);
\draw[-, red] (1.5,1.5) -- (3,1.5) node[xshift=0.3cm] {$\pi_H$};
\draw[dashed, blue] (0,1.35) -- (3,1.35) node[xshift=0.3cm] {$\pi_A$};
\draw [decorate,decoration={brace,amplitude=8pt},xshift=-0.01cm,yshift=0pt]
(1.5,0.8) -- (1.5,1.5) node [black,midway,xshift=-0.8cm]
{\footnotesize $\alpha(\pi_H)$};
\end{tikzpicture}
\caption{Compliance-robust fairness}
\label{fig:example:never_reduce_fairness}
\end{subfigure}
\caption{Examples with trivial individual types (i.e., $\mathcal{X}=\{1\}$) with the same human policy $\pi_H$ and two different algorithmic policies $\pi_A$. The human policy is unfair ($\overline{\pi}_H(0) \neq \overline{\pi}_H(1)$), but the algorithmic policy in both cases is fair in isolation ($\overline{\pi}_A(0) = \overline{\pi}_A(1)$). Left: If the human selectively complies when $a=1$, fairness is reduced (relative to $\pi_H$). Right: Fairness is never reduced for any compliance $c$, i.e., $\pi_A$ is compliance-robustly fair.}
\label{fig:example}
\end{figure}

Our first main result characterizes the class of compliance-robust policies $\Pi_{\text{fair}}$. For intuition, consider the simple example depicted in Figure \ref{fig:example}, where there are no types (i.e., $\mathcal{X}=\{1\}$). The left and right panels consider the same unfair human policy $\pi_H$ (i.e., $\overline{\pi}_H(0) \neq \overline{\pi}_H(1)$) but two different traditionally fair algorithmic policies $\pi_A$ (i.e., $\overline{\pi}_A(0) = \overline{\pi}_A(1)$). On the left, if the human selectively complies when $a=1$, fairness reduces relative to $\pi_H$, i.e., $\pi_A$ is not compliance-robustly fair although it is fair in isolation. On the right, it is easy to check that no compliance function reduces fairness, i.e., $\pi_A$ is compliance-robustly fair. In general, as we formalize next, compliance-robustness holds exactly when $\pi_A$ is ``sandwiched'' between $\pi_H(x,0)$ and $\pi_H(x,1)$.
\begin{restatable}{theorem}{thmMain}
\label{thm:main}
Given $\pi_H$, an algorithmic policy $\pi_A$ is compliance-robustly fair if and only if
\begin{align}
\label{eqn:keycond1}
\alpha(\pi_A)
&\le \alpha(\pi_H) \\
\label{eqn:keycond2}
\pi_H(x,0) &\le \pi_A(x,0)
\qquad(\forall x\in\mathcal{X}) \\
\label{eqn:keycond3}
\pi_A(x,1) &\le \pi_H(x,1)
\qquad(\forall x\in\mathcal{X}).
\end{align}
\end{restatable}
We give a proof in Appendix \ref{appendix:proof_sec3}. In general, it is easier to find compliance-robustly fair policies when the human policy is rather unfair. In fact, the following corollary (proof in Appendix \ref{appendix:proof_sec3}) shows that when the human policy is perfectly fair, there are no nontrivial compliance-robust policies. This is because any deviation by the human that unequally affects the two classes $a\in \mathcal{A}$ provably results in unfairness.

\begin{restatable}{corollary}{corFairHuman}
\label{cor:fair-human}
If $\alpha(\pi_H) = 0$, then $\pi_A$ is compliance-robustly fair if and only if $\pi_A(x,a) = \pi_H(x,a)$ for all $x\in\mathcal{X}$ and $a\in\mathcal{A}$.
\end{restatable}

Theorem~\ref{thm:main} allows us to write down a simple optimization problem (see Algorithm~\ref{fig:algorithm}) to compute a compliance-robustly fair policy $\pi_0$ that performs the best (i.e., minimizes the loss $L$). Note that we have assumed nothing on the class of compliance functions $c(x,a)$ thus far. If we have a priori knowledge that $c$ only depends on some subset of the type variables, then we can trivially remove the constraints in Algorithm~\ref{fig:algorithm} corresponding to those types, enlarging the class of compliance-robust policies $\Pi_{\text{fair}}$.

\begin{algorithm}
\KwIn{Human policy $\pi_H$}
Solve the following optimization problem:
\begin{align*}
    \pi_0 &= \operatorname*{\arg\min}_\pi L(\pi) \\
    \text{subj. to} \quad & \alpha(\pi) \leq \alpha(\pi_H), \\
                          & \pi_H(x,0) \leq \pi(x,0), \quad \forall x \in \mathcal{X}, \\
                          & \pi(x,1) \leq \pi_H(x,1), \quad \forall x \in \mathcal{X}.
\end{align*}
\KwRet{policy $\pi_0$}
\caption{Compliance-robustly Fair Algorithm.}
\label{fig:algorithm}
\end{algorithm}

\paragraph{Extensions.} As noted earlier, it can be easily shown that Theorem~\ref{thm:main} holds for a more general class of compliance functions. First, in practice, compliance may depend on the output of $\pi_A$---e.g., the human decision-maker may comply only when the recommendation $\pi_A(x,a)$ is sufficiently close to their own judgment $\pi_H(x,a)$. In particular, consider a policy-dependent compliance function which depends not only on the type and protected attribute, but also on the output of $\pi_A$, i.e., $\tilde{c}: \mathcal{X}\times\mathcal{A}\times [0,1] \mapsto \{0,1\}$. Then, since $\pi_A$ is itself a function of $x$ and $a$, there exists some compliance function from our original class such that
\begin{align*}
c(x,a)=\tilde{c}(x,a,\pi_A(x,a)).
\end{align*}
Thus, Theorem~\ref{thm:main} automatically subsumes this case.

Second, in practice, human behavior shows significant randomness. Our results hold when the compliance function is stochastic rather than deterministic. Specifically, consider a random compliance function of the form:
\begin{align*}
c_p(x,a) = \begin{cases}
1 & \text{with probability } p(x,a) \\
0 & \text{otherwise},
\end{cases}
\end{align*}
yielding the joint human-AI policy
\begin{align}
\label{pi_C_prob}
\pi_C(x,a) &= \pi_A(x,a)\cdot p(x,a) \\
&\qquad + \pi_H(x,a)\cdot(1-p(x,a)). \nonumber
\end{align}
The proof of Theorem~\ref{thm:main} works without modification for this class.

Next, our results hold for partial compliance functions. Specifically, consider compliance functions of the form $c: \mathcal{X} \times \mathcal{A} \rightarrow [0,1]$, and consider the corresponding human-AI policy $\pi_C$:
\begin{align}
\label{pi_C_partial}
\pi_C(x,a) &= \pi_A(x,a) \cdot c(x,a) + \\
&\qquad \pi_H(x,a) \cdot (1 - c(x,a)). \nonumber
\end{align}
This compliance function allows humans to take a weighted average of their own prediction and the AI prediction when making a final decision.  Our approach works for partial compliance functions for the same reason it works for probabilistic compliance---indeed, note that (\ref{pi_C_partial}) is equivalent to (\ref{pi_C_prob}), so our compliance-robust policy remains robust to partial compliance functions. This form of compliance is well-supported by existing empirical evidence on human behavior, e.g., \citet{balakrishnan2025human} show that human decision-makers often take a constant weighted average between the algorithm’s recommendation and the prediction they would have made independently---a behavior they term ``na\"{i}ve advice-weighting''.


\paragraph{Noisy Estimation of $\pi_H$.} Another issue that arises in practice is that we often do not directly observe $\pi_H$; rather, one must estimate $\hat\pi_H\approx\pi_H$ using supervised learning on historical data prior to the algorithmic intervention (we illustrate this on court sentencing data in Section \ref{ssec:exp-setup}). When using $\hat\pi_H$ instead of $\pi_H$, our compliance-robustness guarantee gracefully degrades in the estimation error of $\hat\pi_H$ as follows. In particular, suppose that we have an estimate $\hat{\pi}_H$ of $\pi_H$ satisfying $|\hat{\pi}_H(x,a)-\pi_H(x,a)|\le\epsilon$ for all $x\in\mathcal{X}$ and $a\in\mathcal{A}$. If we run our algorithm using $\hat\pi_H$, then we obtain an algorithmic policy $\pi_A$ that is compliance-robustly fair for $\hat{\pi}_H$. Now, let $\hat{\pi}_C$ be the joint policy combining $\pi_A$ and $\hat{\pi}_H$, and let $\pi_C$ be the joint policy combining $\pi_A$ and $\pi_H$. Note that for any compliance function $c$, we have $|\hat{\pi}_C(x,a)-\pi_C(x,a)|\le\epsilon$, from which it follows that $\alpha(\pi_C)\le\alpha(\hat{\pi}_C)+2\epsilon$. As a consequence, we have
\begin{align*}
\alpha(\pi_C)
\le \alpha(\hat{\pi}_C) + 2\epsilon
\le \alpha(\hat\pi_H) + 2\epsilon
\le \alpha(\pi_H) + 4\epsilon,
\end{align*}
where the second inequality follows from compliance-robust fairness of $\pi_A$ to $\hat\pi_H$, and the third follows since $\alpha(\hat{\pi}_H)\le\alpha(\pi_H)+2\epsilon$ by our assumption on the estimation error of $\hat\pi_H$. Thus, $\hat{\pi}_A$ satisfies a compliance-robust fairness guarantee within a slack of $4\epsilon$.

\section{Performance of Compliance-Robustly Fair Policies}
\label{section:performance}

We have so far established the best-performing compliance-robust policy $\pi_0$ (defined in Algorithm~\ref{fig:algorithm}) as a strong candidate for algorithmic advice. However, in most cases, we would only provide algorithmic advice if we think it may perform better than the current human policy, i.e., $L(\pi_0) < L(\pi_H)$. In this section, we provide simple conditions (that can be easily verified with knowledge of $\pi_H$ and the performance-maximizing optimal $\pi_*$) to see if algorithmic advice is desirable.

Given the human policy $\pi_H$ and the optimal policy $\pi_*$, let
\begin{align*}
u(a) &= \{x\in\mathcal{X}\mid \pi_H(x,a) \geq \pi_*(x,a)\} \\
\ell(a) &= \{x\in\mathcal{X}\mid \pi_H(x,a) < \pi_*(x,a)\}.
\end{align*}
Intuitively, $u(a)$ denotes regions within group $a$ where the human policy $\pi_H$ assigns weakly higher scores than the optimal policy $\pi_*$. Conversely, $\ell(a)$ corresponds to the regions where $\pi_H$  assigns scores that are strictly lower than those assigned by $\pi_*$.

We now construct the following policy $\pi_B$, which attempts to bridge between achieving high performance  and ensuring compliance-robustness with respect to $\pi_H$:
\begin{align*}
\pi_B(x,a)
&=\begin{cases}
\pi_H(x,a)&\text{if }x\in u(0)\cup\ell(1) \\
\pi_*(x,a)&\text{otherwise}.
\end{cases}
\end{align*}
This policy attempts to maximize performance by matching $\pi_*$ while satisfying constraints (2)-(3) in Theorem~\ref{thm:main} to ensure compliance-robustness. To provide some intuition, $\ell(1)$ represents the regions where $\pi_H$ assigns lower scores than $\pi_*$ for the advantageous group. As stated in Assumption~\ref{assump:loss}, we can improve the performance of $\pi_H$ by increasing $\pi_H$'s scores in $\ell(1)$. However, Theorem~\ref{thm:main} shows that compliance-robustly fair policies cannot return higher scores for any type within the advantageous group. Therefore, the best-performing compliance-robustly fair policies must be the same as $\pi_H$ in $\ell(1)$. The same argument holds for $u(0)$. 

The policy $\pi_B$ will be pivotal for us to understand the performance of compliance-robustly fair policies, as well as their relationship to traditional fairness (in the next section). First, $\pi_B$ is compliance-robustly fair if it (in isolation) does not reduce fairness relative to $\pi_H$ (Lemma~\ref{lem:pib2} in Appendix \ref{appendix:proof_sec4}). Consequently, $\pi_B$ provides a constructive upper bound on the performance of any compliance-robustly fair policy (Lemma~\ref{lem:pib1}), which will be useful for examining when the performance of $\pi_0$ exceeds that of $\pi_H$. Intuitively, if there exists a compliance-robustly fair policy that can improve human performance, then $\pi_B$ must be more accurate than $\pi_H$.

Our next result, Theorem~\ref{thm:existence}, shows simple conditions under which the optimal compliance-robustly fair policy $\pi_0$ is worth sharing with the human (i.e., when $L(\pi_0) < L(\pi_H)$). Namely, we require that human policy $\pi_H$ is not perfectly fair (in which case, there is no nontrivial compliance-robustly fair policy by Corollary~\ref{cor:fair-human}), and $\pi_H$ deviates from the performance-optimal policy $\pi_*$ in a direction that we can plausibly correct with algorithmic advice.

\begin{restatable}{theorem}{thmExistence}
\label{thm:existence}
Assume that $\alpha(\pi_H)\neq0$, and that either $\pi_H(x,1)\neq\pi_*(x,1)$ for some $x\in u(1)$ or $\pi_H(x,0)\neq\pi_*(x,0)$ for some $x\in \ell(0)$. Then, we have $L(\pi_0)<L(\pi_H)$.
\end{restatable}
We give a proof in Appendix~\ref{appendix:proof_sec4}. As discussed earlier, $\pi_B$ must equal $\pi_H$ in $u(0)$ and $\ell(1)$. Consequently, compliance-robustly fair policies can only perform better than $\pi_H$ in $\ell(0)$ and $u(1)$. As long as the human policy $\pi_H$ doesn't perfectly match the optimal policy $\pi_*$ in at least one of these regions, we can construct a compliance-robustly fair policy that achieves strictly better performance than $\pi_H$.

\section{Compliance-Robust Fairness vs. Traditional Fairness}
\label{section:traditional_fairness}
As shown in the last section, compliance-robust fairness and performance improvement are often compatible; the same holds for traditional fairness and performance improvement \cite{hardt2016equality}. However, we will show that there is considerable tension between maintaining \textit{both} types of fairness (compliance-robust fairness and traditional algorithmic fairness) while improving performance.

Building on the mild conditions required for a performance-improving compliance-robust policy in Theorem~\ref{thm:existence}, the next lemma establishes additional conditions that are necessary and sufficient to find a policy $\pi_A$ that is also traditionally fair (i.e., $\alpha(\pi_A) = 0$).

\begin{restatable}{lemma}{lemTraditional}
\label{lem:traditional}
Assume that $\alpha(\pi_H)\neq0$, and that either $\pi_H(x,1)\neq\pi_*(x,1)$ for some $x \in u(1)$ or $\pi_H(x,0)\neq\pi_*(x,0)$ for some $x\in \ell(0)$. Then, there exists a compliance-robustly fair policy $\pi_A$ that is also traditionally fair ($\alpha(\pi_A)=0$) and performance-improving ($L(\pi_A)<L(\pi_H)$) if and only if there exists a policy $\pi$ satisfying
\begin{align}
\overline{\pi}(1) &\leq \overline{\pi}(0) 
\label{eqn:traditional:1} \\
\pi(x,1) &\leq \pi_B(x,1)
\qquad\qquad(\forall x\in\mathcal{X})
\label{eqn:traditional:2} \\
\pi(x,0) &\geq \pi_B(x,0)
\qquad\qquad(\forall x\in\mathcal{X})
\label{eqn:traditional:3} \\
L(\pi) &<L(\pi_H).
\label{eqn:traditional:4}
\end{align}
\end{restatable}
We give a proof in Appendix~\ref{appendix:proof_sec5}.
Next, we show a natural setting where we meet the above conditions---namely, when the data-generating process is such that the optimal performance-maximizing policy $\pi_*$ is already perfectly fair without any added constraints (i.e., $\alpha(\pi_*)=0$).

\begin{theorem}
Assume that $\alpha(\pi_H)\neq0$, and that either $\pi_H(x,1)\neq\pi_*(x,1)$ for some $x \in u(1)$ or $\pi_H(x,0)\neq\pi_*(x,0)$ for some $x \in \ell(0)$. Then, if $\pi_*$ is fair, there is always a compliance-robustly fair $\pi_A \in \Pi_{\text{fair}}$ that is also traditionally fair and performance-improving.
\end{theorem}
\begin{customproof}
Consider $\pi_B$. Since $\pi_*$ is fair and $\pi_B(x,1) \leq \pi_*(x,1)$ and $\pi_B(x,0) \geq \pi_*(x,0)$, it immediately follows that $\overline{\pi}_B(1) \leq \overline{\pi}_B(0)$. Thus, the claim follows from Lemma~\ref{lem:traditional} (with $\pi = \pi_B$).
\end{customproof}


Unfortunately, it unlikely that an unconstrained performance-maximizing policy will be inherently fair; this insight has been the driving force of the algorithmic fairness literature. Rather, we may have to choose between the properties of compliance-robust fairness (to avoid disparate harm relative to the human policy), performance improvement (to ensure that algorithmic recommendations actually drive improved decisions), and traditional fairness (to ensure the algorithm is fair in isolation). To this end, we now construct a simple setting where we can only satisfy one criterion---performance improvement \textit{or} traditional fairness---for all compliance-robustly fair policies.

Intuitively, this tension can arise when the human policy is not far from the performance-maximizing policy ($\pi_H \approx \pi_*$) and this policy is quite unfair ($\alpha(\pi_*) \gg 0$). Consider the extreme case where $\pi_H = \pi_*$ and $\alpha(\pi_H) > 0$. By Theorem~\ref{thm:main}, $\pi_*$ is compliance-robustly fair, and yet it is not traditionally fair. Thus, any traditionally fair policy must necessarily perform worse than the existing $\pi_H$ or not be compliance-robustly fair. The following proposition crystallizes this intuition in a nontrivial setting.


\begin{restatable}{proposition}{propositionNoneExist}
\label{prop:non_exist}
There exists $\mathcal{X}$, $\mathbb{P}$, $L$, and $\pi_H$ satisfying $\alpha(\pi_H)\neq0$ and $\pi_H\neq\pi^*$
such that for any policy $\pi$, $\pi$ cannot simultaneously satisfy all of the following: (i) $\pi\in\Pi_{\text{fair}}$, (ii) $\alpha(\pi)=0$, and (iii) $L(\pi)\le L(\pi_H)$.
\end{restatable}
We give a proof in Appendix~\ref{appendix:proof_sec5}. Given these results, if the goal is to improve fairness and accuracy in human-AI collaboration outcomes, it may be preferable to design an algorithmic policy that is accurate and compliance-robustly fair, but not fair in isolation.

One may question whether the challenges arising from selective compliance and the resulting trade-offs are only relevant to our fairness definition---equality of opportunity \citep{hardt2016equality}. Therefore, we show in Appendix \ref{general_fairness} that selective compliance can lead to undesirable outcomes for a large class of fairness definitions that satisfies a mild assumption.

\section{Empirical Evaluation}

We empirically simulate the performance of our compliance-robustly fair algorithm using criminal sentencing data from Virginia from 2000 to 2004. In July 2002, the Virginia Criminal Sentencing Commission (VCSC) introduced an algorithmic risk assessment tool to help judges identify \textit{low-risk} individuals with a felony conviction, with the goal of diverting them from prison. Before making final sentencing decisions, judges were presented with the model's predicted risk score to facilitate risk assessment. We leverage data pre- and post- introduction of the risk assessment tool to assess the fairness and performance of different algorithmic advice policies.


\subsection{Experimental Setup} \label{ssec:exp-setup}


\paragraph{Data.} Following \citet{stevenson2022algorithmic}, we obtained criminal sentencing records through a Freedom of Information Act request, which we merged with defendant demographics from \url{https://virginiacourtdata.org/}. This data spans $22,433$ sentencing events made by 206 different judges (see Appendix~\ref{exp-details}). We use data prior to the launch of the tool ($N=15,106$) to estimate each judge's policy $\pi_H$, and data from 2003 and 2004 ($N=7,327$) for evaluation. 


The true outcome $y$ denotes whether a defendant recidivates within three years following release.\footnote{In practice, we must also address the issue of \textit{selective labels}---we only observe the true outcome when a defendant is released~\citep{lakkaraju2017selective}. However, this issue only marginally affects our results, since we only examine individuals that are eligible for the risk assessment tool---more than 99\% of eligible defendants in our sample were released before 2010, so there is negligible censoring of observed outcomes. In particular, we observe the outcome of interest, three-year recidivism, for nearly every case in our dataset.} All decision-making policies, including $\pi_H$ and $\pi_A$, output risk scores representing the estimated probability that a defendant is low risk (i.e., unlikely to recidivate) based on the observed defendant features. Defendants with higher predicted risk scores are more likely to receive reduced sentences. The protected attribute $a$ is race, restricted to either White or Black.

\paragraph{Estimating the judges' policies.}

To construct $\pi_H$, we need the judges' \textit{independent} assessment of whether a defendant should be offered a reduced sentence. We estimate this by examining when judges overrode pre-existing VCSC sentencing guidelines to reduce a defendant's sentence (see Appendix~\ref{exp-details} for details), prior to the introduction of the algorithmic risk assessment tool. We train a gradient boosted decision tree~\citep{ke2017lightgbm} to predict reduced sentences based on observed defendant covariates as well as the the Judge ID (to obtain judge-specific policies).

\paragraph{Estimating the judges' compliance functions.} After the introduction of the algorithmic risk assessment tool, compliance with the tool's recommendations is an observed variable in the data. We estimate a judge's compliance function $c$ by training a gradient boosted decision tree to predict compliance using the same set of defendant covariates as above.

\paragraph{Estimating the original risk assessment model.}

We do not have access to the original VCSC risk assessment tool, but we observe the tool's recommendations (i.e., low-risk or not). Thus, we train a gradient boosted decision tree to predict the tool's policy $\pi_A^{\text{actual}}$, using the same set of defendant covariates as above (except for Judge ID).

\paragraph{Policies.} We then construct different human-AI collaborative policies for each judge using our estimates of judge-specific $\pi_H, c$ and $\pi_A$---(i) the actual observed policy $\pi_C^{\text{actual}}(x,a)$, (ii) our compliance-robustly fair policy $\pi^{\text{robust}}_C(x,a)$,\footnote{Using the estimated $\pi_H$, we solve the optimization problem in Algorithm~\ref{fig:algorithm} to construct the compliance-robustly fair policy $\pi_A^{\text{robust}}$ for each judge. In particular, we represent $\pi_A^{\text{robust}}$ as a lookup table. If a new $(x',a')$ is encountered during test time (i.e., $(x',a')$ is not present in the lookup table), we define $\pi_A^{\text{robust}}(x',a')$ to be $\pi_H(x',a')$.} (iii) the performance-maximizing policy $\pi_C^*(x,a)$, and (iv) the traditionally fair policy $\pi_C^{\text{trad-fair}}(x,a)$. Note that these are all human-AI policies and may not satisfy the properties guaranteed by their respective algorithmic policies $\pi_A$ alone.
When simulating the performance and fairness of $\pi_A^{\text{robust}}$, $\pi_A^{*}$ and $\pi_A^{\text{trad-fair}}$, we make a key assumption that judges' compliance functions would remain the same for these alternative algorithmic risk assessment tools as in the original VCSC algorithmic risk assessment tool. This may not be the case in practice, but our compliance-robust approach guarantees hold under \textit{any} new compliance function that judges may adopt.

Note that there may be errors in our estimated policies since the information in our dataset may not exactly match the information available to judges at the time of decision-making. However, we expect our simulation results to remain informative when comparing against other policies that are also trained on the same available data.

\paragraph{Metrics.}
For a human-AI policy $\pi_C$, we examine both performance improvement, $L(\pi_H) - L(\pi_C)$, and fairness improvement, $\alpha(\pi_H) - \alpha(\pi_C)$; details in Appendix~\ref{exp-details}. 

\begin{figure}[ht]
\centering
\begin{subfigure}{0.4\textwidth}
\centering
\includegraphics[width=\linewidth]{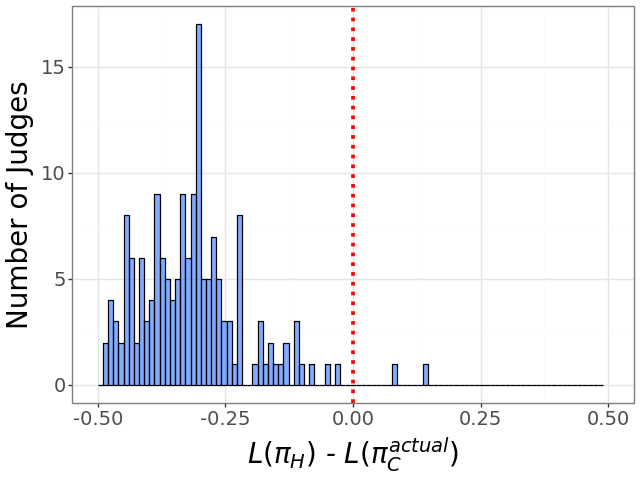} 
\caption{Performance of $\pi_C^{actual}$}
\label{fig:actual_loss_diff}
\end{subfigure}
\hfill
\begin{subfigure}{0.4\textwidth}
\centering
\includegraphics[width=\linewidth]{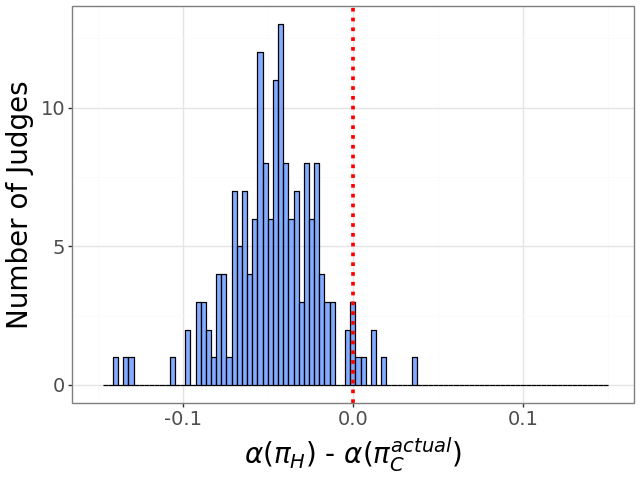} 
\caption{Fairness of $\pi_C^{actual}$}
\label{fig:actual_fairness_diff}
\end{subfigure}
\\
\begin{subfigure}{0.4\textwidth}
\centering
\includegraphics[width=\linewidth]{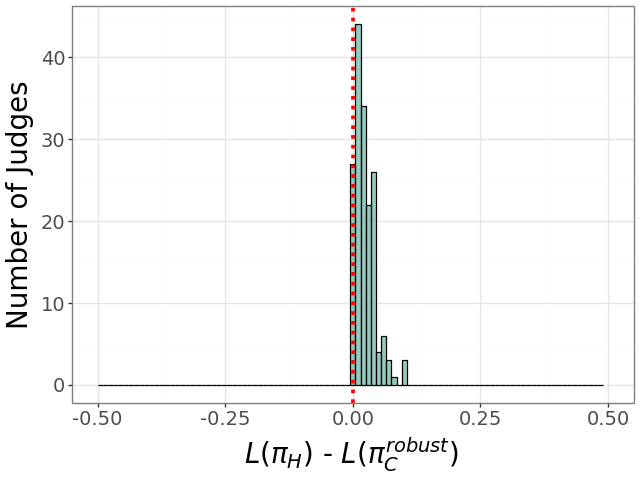} 
\caption{Performance of $\pi_C^{robust}$}
\label{fig:robust_loss_diff}
\end{subfigure}
\hfill
\begin{subfigure}{0.4\textwidth}
\centering
\includegraphics[width=\linewidth]{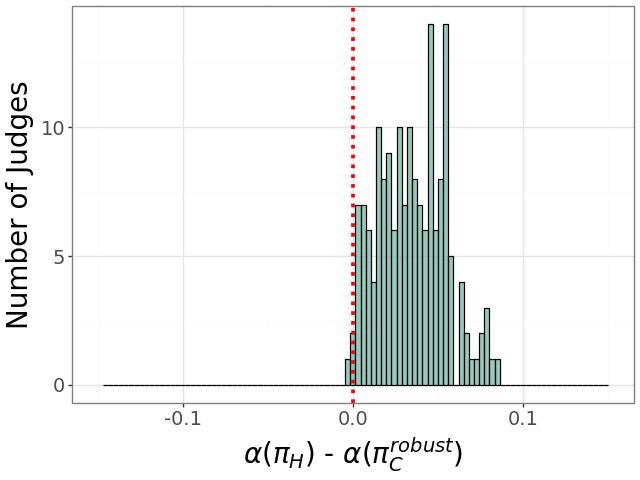} 
\caption{Fairness of $\pi_C^{robust}$}
\label{fig:robust_fairness_diff}
\end{subfigure}
\\
\begin{subfigure}[b]{0.4\textwidth}
\centering
\includegraphics[width=\textwidth]{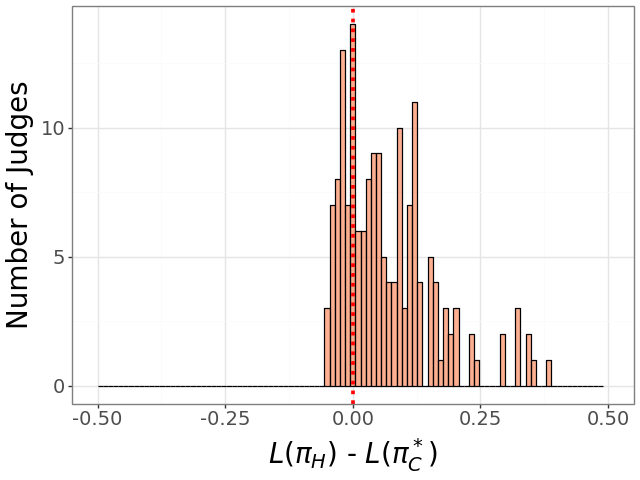}
\caption{Performance of $\pi_C^{*}$}
\label{fig:star_comparison_acc}
\end{subfigure}
\hfill
\begin{subfigure}[b]{0.4\textwidth}
\centering
\includegraphics[width=\textwidth]{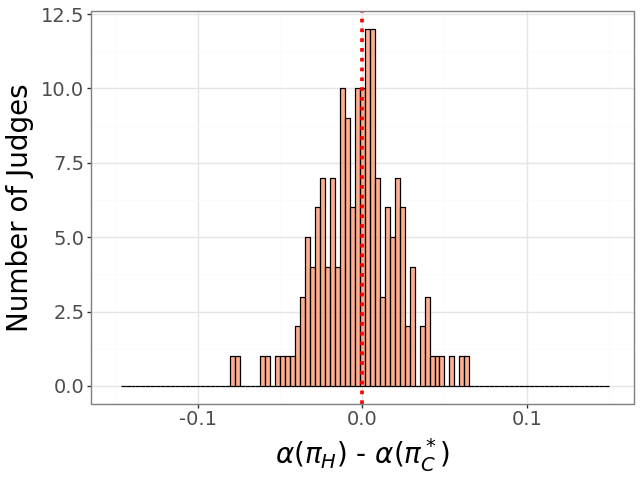}
\caption{Fairness of $\pi_C^{*}$}
\label{fig:star_comparison_fairness}
\end{subfigure}
\begin{subfigure}[b]{0.4\textwidth}
\centering
\includegraphics[width=\textwidth]{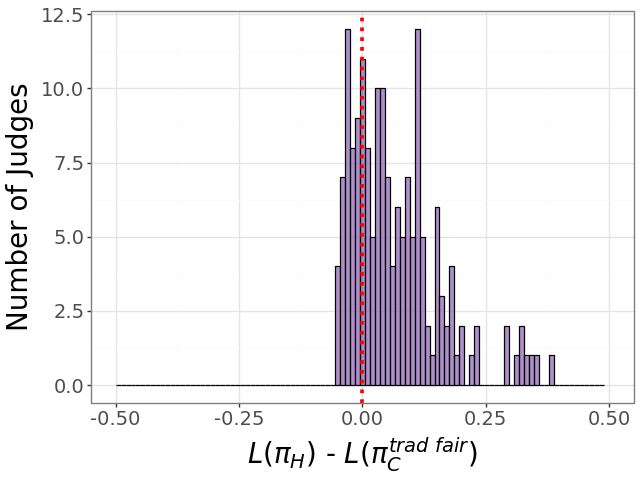}
\caption{Performance of $\pi_C^{\text{trad-fair}}$}
\label{fig:fair_comparison_acc}
\end{subfigure}
\hfill
\begin{subfigure}[b]{0.4\textwidth}
\centering
\includegraphics[width=\textwidth]{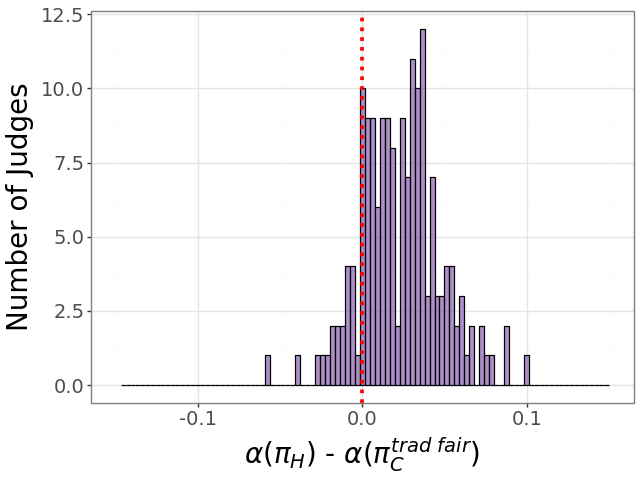}
\caption{Fairness of $\pi_C^{\text{trad-fair}}$}
\label{fig:fair_comparison_fairness}
\end{subfigure}

\caption{\scriptsize We show the performance and fairness comparisons for $\pi_C^{\text{actual}}$, $\pi_C^{\text{robust}}$, $\pi_C^*$ and $\pi_C^{\text{trad-fair}}$ across the 170 judges in our evaluation sample. Bars to the right of the red dotted line correspond to judges whose accuracy or fairness improve with the algorithmic recommendation. }
\label{fig:judge_level_analyses}
\end{figure}

\subsection{Results} \label{exp-results}

Figure~\ref{fig:judge_level_analyses} shows a judge-level comparison of each of the four human-AI policies (relative to the $\pi_H$) in terms of performance and fairness. First, as discussed in the findings of \citet{stevenson2022algorithmic}, we observe that the actual VCSC reduced performance (Fig~\ref{fig:actual_loss_diff}) and fairness (Fig~\ref{fig:actual_fairness_diff}), relative to the prior human-alone policy, for nearly every judge. In contrast, our compliance-robust policy $\pi_C^{\text{robust}}$ benefits almost every judge in terms of both performance (Fig~\ref{fig:robust_loss_diff}) and fairness (Fig~\ref{fig:robust_fairness_diff}). Only 2 of 170 judges have a negligible deterioration in fairness, likely due to finite sample estimation error. Then, as expected, the policy $\pi_C^*$ that relies on a performance maximizing algorithm  significantly improves performance (Fig~\ref{fig:star_comparison_acc}), but comes at the cost of 54\% of judges seeing deterioration in fairness in their sentencing outcomes (Fig~\ref{fig:star_comparison_fairness}). Finally, we consider the policy $\pi_C^{\text{trad-fair}}$ that relies on the highest-performing traditionally fair algorithm---we find that while it improves accuracy (Fig~\ref{fig:fair_comparison_acc}) and fairness (Fig~\ref{fig:fair_comparison_fairness}) \textit{on average}, 27\% of judges see reduced performance and 14\% of judges see deterioration in fairness due to selective compliance. In contrast, our compliance-robust approach guarantees weakly improved performance and fairness for \textit{every} judge, regardless of their compliance pattern.

\paragraph{Mechanism.} As illustrated in Figure~\ref{fig:example}, algorithmic recommendations can reduce fairness when decision-makers disproportionately comply with the algorithmic recommendations for an advantaged group whenever the algorithm offers a more favorable decision. To shed more light, we examine the compliance pattern $c_{\text{problem}}$ and human-alone policies $\pi^{\text{Ave}}_H$ for the subset of judges that worsen fairness the most (see Appendix~\ref{exp-details} for details). We define the variable ``AI Low Risk'' for a defendant $i$ with features $(x_i, a_i)$ as the indicator function of whether the algorithmic policy is more lenient than the human alone policy, $\pi_A(x_i, a_i) > \pi_H(x_i, a_i)$. Then, we test if judges comply more frequently for White defendants when the algorithmic policy is more lenient:
\begin{align*}
    &P(\text{Comply}_i=1) =\\
    &\quad \text{Logit}(\beta_0 + \beta_1 \cdot \text{White}_i \\
    &\quad\quad + \beta_2 \cdot \text{AI Low Risk}_i \\
    &\quad\quad + \beta_3 \cdot (\text{AI Low Risk}_i \times \text{White}_i) + \epsilon_i).
\end{align*}
Indeed, we find that $\beta_3$ is positive and statistically significant for all human-AI collaborative policies except our compliance-robust policy, indicating that judges' compliance behaviors exacerbate existing racial biases under these policies. In contrast, our compliance-robustly fair policy ($\pi_A^{\text{robust}}$) effectively guards against such problematic compliance behaviors.

\section{Conclusion}
This paper illustrates the perils of selective compliance for equitable outcomes in human-AI collaboration. In particular, even algorithms that satisfy traditional algorithmic fairness criteria can amplify unfairness in decisions (relative to the human making decisions in isolation). Unfortunately, a human decision-maker's compliance pattern is a priori unknown, and may even change over time, affecting fairness in outcomes. Therefore, we introduce the concept of compliance-robust fairness and demonstrate how to derive algorithmic policies that weakly improve fairness regardless of the human's compliance pattern. Naturally, it is also important that the algorithmic advice achieves better performance than the human alone. We show that, as long as the human policy is slightly sub-optimal and not perfectly fair, the best performance-improving compliance-robust policy still generates improvements over the human in isolation. However, it is not always the case that we can also achieve the third property of traditional fairness---we may need to rely on algorithmic policies that are unfair in isolation to achieve compliance-robustly fair human-AI collaboration. We illustrate our approach on criminal sentencing data from Virginia. We demonstrate significant gains in fairness compared to a traditionally fair policy that does not account for judges' selective compliance patterns. Our findings contribute to the design of human-AI collaboration systems that are ``user-aware,'' enhancing rather than diminishing fairness in collaborative decisions.

\clearpage


%
%
%



\normalsize
\bibliographystyle{informs2014} 
\bibliography{ref} 

\begin{thebibliography}{39}
\providecommand{\natexlab}[1]{#1}
\providecommand{\url}[1]{\texttt{#1}}
\providecommand{\urlprefix}{URL }

\bibitem[{Agarwal et~al.(2023)Agarwal, Moehring, Rajpurkar, \protect\BIBand{}
  Salz}]{agarwal2023combining}
Agarwal N, Moehring A, Rajpurkar P, Salz T (2023) Combining human expertise
  with artificial intelligence: Experimental evidence from radiology. Technical
  report, National Bureau of Economic Research.

\bibitem[{Ahn et~al.(2024)Ahn, Almaatouq, Gulabani, \protect\BIBand{}
  Hosanagar}]{ahn2024impact}
Ahn D, Almaatouq A, Gulabani M, Hosanagar K (2024) Impact of model
  interpretability and outcome feedback on trust in ai. \emph{Proceedings of
  the CHI Conference on Human Factors in Computing Systems}, 1--25.

\bibitem[{Albright et~al.(2019)}]{albright2019if}
Albright A, et~al. (2019) If you give a judge a risk score: evidence from
  kentucky bail decisions. \emph{Law, Economics, and Business Fellows’
  Discussion Paper Series} 85:2019--1.

\bibitem[{Alur et~al.(2024)Alur, Raghavan, \protect\BIBand{}
  Shah}]{alur2024human}
Alur R, Raghavan M, Shah D (2024) Human expertise in algorithmic prediction.
  \emph{Advances in Neural Information Processing Systems} 37:138088--138129.

\bibitem[{Bai et~al.(2022)Bai, Dai, Zhang, Zhang, \protect\BIBand{}
  Hu}]{bai2022impacts}
Bai B, Dai H, Zhang DJ, Zhang F, Hu H (2022) The impacts of algorithmic work
  assignment on fairness perceptions and productivity: Evidence from field
  experiments. \emph{Manufacturing \& Service Operations Management}
  24(6):3060--3078.

\bibitem[{Balakrishnan et~al.(2025)Balakrishnan, Ferreira, \protect\BIBand{}
  Tong}]{balakrishnan2025human}
Balakrishnan M, Ferreira KJ, Tong J (2025) Human-algorithm collaboration with
  private information: Na{\"\i}ve advice-weighting behavior and mitigation.
  \emph{Management Science} .

\bibitem[{Bastani et~al.(2021)Bastani, Bastani, \protect\BIBand{}
  Sinchaisri}]{bastani2021improving}
Bastani H, Bastani O, Sinchaisri WP (2021) Improving human decision-making with
  machine learning. \emph{arXiv preprint arXiv:2108.08454} .

\bibitem[{Bastani et~al.(2022)Bastani, Gupta, Jung, Noarov, Ramalingam,
  \protect\BIBand{} Roth}]{bastani2022practical}
Bastani O, Gupta V, Jung C, Noarov G, Ramalingam R, Roth A (2022) Practical
  adversarial multivalid conformal prediction. \emph{Advances in Neural
  Information Processing Systems} 35:29362--29373.

\bibitem[{Basu(2023)}]{basu2023use}
Basu A (2023) Use of race in clinical algorithms. \emph{Science Advances}
  9(21):eadd2704.

\bibitem[{Cai et~al.(2020)Cai, Gaebler, Garg, \protect\BIBand{}
  Goel}]{cai2020fair}
Cai W, Gaebler J, Garg N, Goel S (2020) Fair allocation through selective
  information acquisition. \emph{Proceedings of the AAAI/ACM Conference on AI,
  Ethics, and Society}, 22--28.

\bibitem[{Calders et~al.(2009)Calders, Kamiran, \protect\BIBand{}
  Pechenizkiy}]{calders2009building}
Calders T, Kamiran F, Pechenizkiy M (2009) Building classifiers with
  independency constraints. \emph{2009 IEEE international conference on data
  mining workshops}, 13--18 (IEEE).

\bibitem[{Campero et~al.(2022)Campero, Vaccaro, Song, Wen, Almaatouq,
  \protect\BIBand{} Malone}]{campero2022test}
Campero A, Vaccaro M, Song J, Wen H, Almaatouq A, Malone TW (2022) A test for
  evaluating performance in human-computer systems. \emph{arXiv preprint
  arXiv:2206.12390} .

\bibitem[{Chen et~al.(2023)Chen, Wang, Williamson, Chen, Lipkova, Lu, Sahai,
  \protect\BIBand{} Mahmood}]{chen2023algorithmic}
Chen RJ, Wang JJ, Williamson DF, Chen TY, Lipkova J, Lu MY, Sahai S, Mahmood F
  (2023) Algorithmic fairness in artificial intelligence for medicine and
  healthcare. \emph{Nature Biomedical Engineering} 7(6):719--742.

\bibitem[{Corbett-Davies \protect\BIBand{} Goel(2018)}]{corbett2018measure}
Corbett-Davies S, Goel S (2018) The measure and mismeasure of fairness: A
  critical review of fair machine learning. \emph{arXiv preprint
  arXiv:1808.00023} .

\bibitem[{Dwork et~al.(2012)Dwork, Hardt, Pitassi, Reingold, \protect\BIBand{}
  Zemel}]{dwork2012fairness}
Dwork C, Hardt M, Pitassi T, Reingold O, Zemel R (2012) Fairness through
  awareness. \emph{Proceedings of the 3rd innovations in theoretical computer
  science conference}, 214--226.

\bibitem[{Esteva et~al.(2017)Esteva, Kuprel, Novoa, Ko, Swetter, Blau,
  \protect\BIBand{} Thrun}]{esteva2017dermatologist}
Esteva A, Kuprel B, Novoa RA, Ko J, Swetter SM, Blau HM, Thrun S (2017)
  Dermatologist-level classification of skin cancer with deep neural networks.
  \emph{nature} 542(7639):115--118.

\bibitem[{Gillis et~al.(2021)Gillis, McLaughlin, \protect\BIBand{}
  Spiess}]{gillis2021fairness}
Gillis T, McLaughlin B, Spiess J (2021) On the fairness of machine-assisted
  human decisions. \emph{arXiv preprint arXiv:2110.15310} .

\bibitem[{Hardt et~al.(2016)Hardt, Price, \protect\BIBand{}
  Srebro}]{hardt2016equality}
Hardt M, Price E, Srebro N (2016) Equality of opportunity in supervised
  learning. \emph{Advances in neural information processing systems} 29.

\bibitem[{Hoffman et~al.(2018)Hoffman, Kahn, \protect\BIBand{}
  Li}]{hoffman2018discretion}
Hoffman M, Kahn LB, Li D (2018) Discretion in hiring. \emph{The Quarterly
  Journal of Economics} 133(2):765--800.

\bibitem[{Joseph et~al.(2016)Joseph, Kearns, Morgenstern, \protect\BIBand{}
  Roth}]{joseph2016fairness}
Joseph M, Kearns M, Morgenstern JH, Roth A (2016) Fairness in learning: Classic
  and contextual bandits. \emph{Advances in neural information processing
  systems} 29.

\bibitem[{Kallus et~al.(2022)Kallus, Mao, \protect\BIBand{}
  Zhou}]{kallus2022assessing}
Kallus N, Mao X, Zhou A (2022) Assessing algorithmic fairness with unobserved
  protected class using data combination. \emph{Management Science}
  68(3):1959--1981.

\bibitem[{Ke et~al.(2017)Ke, Meng, Finley, Wang, Chen, Ma, Ye,
  \protect\BIBand{} Liu}]{ke2017lightgbm}
Ke G, Meng Q, Finley T, Wang T, Chen W, Ma W, Ye Q, Liu TY (2017) Lightgbm: A
  highly efficient gradient boosting decision tree. \emph{Advances in neural
  information processing systems} 30.

\bibitem[{Kearns et~al.(2019)Kearns, Neel, Roth, \protect\BIBand{}
  Wu}]{kearns2019empirical}
Kearns M, Neel S, Roth A, Wu ZS (2019) An empirical study of rich subgroup
  fairness for machine learning. \emph{Proceedings of the conference on
  fairness, accountability, and transparency}, 100--109.

\bibitem[{Kim et~al.(2019)Kim, Ghorbani, \protect\BIBand{}
  Zou}]{kim2019multiaccuracy}
Kim MP, Ghorbani A, Zou J (2019) Multiaccuracy: Black-box post-processing for
  fairness in classification. \emph{Proceedings of the 2019 AAAI/ACM Conference
  on AI, Ethics, and Society}, 247--254.

\bibitem[{Kleinberg et~al.(2018)Kleinberg, Lakkaraju, Leskovec, Ludwig,
  \protect\BIBand{} Mullainathan}]{kleinberg2018human}
Kleinberg J, Lakkaraju H, Leskovec J, Ludwig J, Mullainathan S (2018) Human
  decisions and machine predictions. \emph{The quarterly journal of economics}
  133(1):237--293.

\bibitem[{Kleinberg et~al.(2016)Kleinberg, Mullainathan, \protect\BIBand{}
  Raghavan}]{kleinberg2016inherent}
Kleinberg J, Mullainathan S, Raghavan M (2016) Inherent trade-offs in the fair
  determination of risk scores. \emph{arXiv preprint arXiv:1609.05807} .

\bibitem[{Lakkaraju et~al.(2017)Lakkaraju, Kleinberg, Leskovec, Ludwig,
  \protect\BIBand{} Mullainathan}]{lakkaraju2017selective}
Lakkaraju H, Kleinberg J, Leskovec J, Ludwig J, Mullainathan S (2017) The
  selective labels problem: Evaluating algorithmic predictions in the presence
  of unobservables. \emph{Proceedings of the 23rd ACM SIGKDD International
  Conference on Knowledge Discovery and Data Mining}, 275--284.

\bibitem[{Manshadi et~al.(2023)Manshadi, Niazadeh, \protect\BIBand{}
  Rodilitz}]{manshadi2023fair}
Manshadi V, Niazadeh R, Rodilitz S (2023) Fair dynamic rationing.
  \emph{Management Science} 69(11):6818--6836.

\bibitem[{Morgan \protect\BIBand{} Pass(2019)}]{morgan2019paradoxes}
Morgan A, Pass R (2019) Paradoxes in fair computer-aided decision making.
  \emph{Proceedings of the 2019 AAAI/ACM Conference on AI, Ethics, and
  Society}, 85--90.

\bibitem[{Mulvany \protect\BIBand{} Randhawa(2021)}]{mulvany2021fair}
Mulvany J, Randhawa RS (2021) Fair scheduling of heterogeneous customer
  populations. \emph{Available at SSRN 3803016} .

\bibitem[{Newman et~al.(2020)Newman, Fast, \protect\BIBand{}
  Harmon}]{newman2020eliminating}
Newman DT, Fast NJ, Harmon DJ (2020) When eliminating bias isn’t fair:
  Algorithmic reductionism and procedural justice in human resource decisions.
  \emph{Organizational Behavior and Human Decision Processes} 160:149--167.

\bibitem[{Rajpurkar et~al.(2022)Rajpurkar, Chen, Banerjee, \protect\BIBand{}
  Topol}]{rajpurkar2022ai}
Rajpurkar P, Chen E, Banerjee O, Topol EJ (2022) Ai in health and medicine.
  \emph{Nature medicine} 28(1):31--38.

\bibitem[{Stevenson \protect\BIBand{} Doleac(2024)}]{stevenson2022algorithmic}
Stevenson MT, Doleac JL (2024) Algorithmic risk assessment in the hands of
  humans. \emph{American Economic Journal, Economic Policy} .

\bibitem[{Tong et~al.(2021)Tong, Jia, Luo, \protect\BIBand{}
  Fang}]{tong2021janus}
Tong S, Jia N, Luo X, Fang Z (2021) The janus face of artificial intelligence
  feedback: Deployment versus disclosure effects on employee performance.
  \emph{Strategic Management Journal} 42(9):1600--1631.

\bibitem[{Van~Dam(2019)}]{van2019algorithms}
Van~Dam A (2019) Algorithms were supposed to make virginia judges fairer. what
  happened was far more complicated. \emph{Available via The Washington Post.
  Retrieved February} 18:2020.

\bibitem[{Wang et~al.(2024)Wang, Gao, \protect\BIBand{}
  Agarwal}]{wang2024friend}
Wang W, Gao G, Agarwal R (2024) Friend or foe? teaming between artificial
  intelligence and workers with variation in experience. \emph{Management
  Science} 70(9):5753--5775.

\bibitem[{Weerts et~al.(2023)Weerts, Dudík, Edgar, Jalali, Lutz,
  \protect\BIBand{} Madaio}]{weerts2023fairlearn}
Weerts H, Dudík M, Edgar R, Jalali A, Lutz R, Madaio M (2023) Fairlearn:
  Assessing and improving fairness of ai systems.
  \urlprefix\url{http://jmlr.org/papers/v24/23-0389.html}.

\bibitem[{Xu \protect\BIBand{} Dean(2023)}]{xu2023decision}
Xu R, Dean S (2023) Decision-aid or controller? steering human decision makers
  with algorithms. \emph{arXiv preprint arXiv:2303.13712} .

\bibitem[{Zliobaite(2015)}]{zliobaite2015relation}
Zliobaite I (2015) On the relation between accuracy and fairness in binary
  classification. \emph{arXiv preprint arXiv:1505.05723} .

\end{thebibliography}

\newpage
\begin{APPENDICES}
\renewcommand{\thesection}{\Alph{section}}

\section{Theoretical Results}
\subsection{Proof of Results in Section \ref{section:main}}
\label{appendix:proof_sec3}
\thmMain*
\begin{customproof}
First, we show that (\ref{eqn:keycond1}), (\ref{eqn:keycond2}), and (\ref{eqn:keycond3}) are sufficient. Note that (\ref{eqn:keycond2}) and (\ref{eqn:keycond3}) imply
\begin{align}
\label{eqn:newcond1}
\overline{\pi}_H(0) &\le \overline{\pi}_C(0) \le \overline{\pi}_A(0) \\
\label{eqn:newcond2}
\overline{\pi}_A(1) &\le \overline{\pi}_C(1) \le \overline{\pi}_H(1)
\end{align}
respectively, for any compliance function $c$. Now, we have
\begin{align*}
\overline{\pi}_C(1) - \overline{\pi}_C(0)
\le \overline{\pi}_H(1) - \overline{\pi}_H(0)
\le \alpha(\pi_H),
\end{align*}
where the first inequality follows from (\ref{eqn:newcond1}) and (\ref{eqn:newcond2}). Additionally, we have
\begin{align*}
\overline{\pi}_C(0) - \overline{\pi}_C(1)
\le \overline{\pi}_A(0) - \overline{\pi}_A(1)
\le \alpha(\pi_A)
\le \alpha(\pi_H).
\end{align*}
where the first inequality follows from (\ref{eqn:newcond1}) and (\ref{eqn:newcond2}), and the third from (\ref{eqn:keycond1}). The claim follows.

Next, we show that (\ref{eqn:keycond1}), (\ref{eqn:keycond2}), and (\ref{eqn:keycond3}) are necessary. Note that (\ref{eqn:keycond1}) is clearly necessary, or the compliance function $c(x,a)=1$ for all $x,a$ (i.e., the human always complies with the algorithmic decision) reduces fairness. To see that (\ref{eqn:keycond2}) is necessary, suppose to the contrary that $\pi_H(0,x_0)>\pi_A(0,x_0)$ for some $x\in\mathcal{X}$. Then, consider the compliance function
\begin{align*}
c(x,a)&=
\begin{cases}
1&\text{if }x=x_0,a=0 \\
0&\text{otherwise}.
\end{cases}
\end{align*}
For this $c$, it is easy to see that by Assumption~\ref{assump:probability}, $\overline{\pi}_H(0)>\overline{\pi}_C(0)$, whereas $\overline{\pi}_C(1)=\overline{\pi}_H(1)$. By Assumption~\ref{assump:group}, it follows that $\alpha(\pi_C)>\alpha(\pi_H)$, so $c$ reduces fairness. The proof for (\ref{eqn:keycond3}) is similar.
\end{customproof}

\corFairHuman*
\begin{customproof}
Consider a compliance-robustly fair policy $\pi_A$, and assume to the contrary that $\pi_A(x_0,a_0)\neq\pi_H(x_0,a_0)$ for some $x_0\in\mathcal{X}$ and $a_0\in\mathcal{A}$. We assume that $a_0=1$; the case $a_0=0$ is similar. By Theorem~\ref{thm:main}, we have $\pi_A(x,1)\le\pi_H(x,1)$ for all $x\in\mathcal{X}$, so $\pi_A(x_0,1)<\pi_H(x_0,1)$. Then, by Assumption~\ref{assump:probability}, we have $\overline{\pi}_A(1)<\overline{\pi}_H(1)$. Also by Theorem~\ref{thm:main}, we have $\pi_A(x,0)\ge\pi_H(x,0)$ for all $x\in\mathcal{X}$, so $\overline{\pi}_A(0)\ge\overline{\pi}_H(0)$. Thus, we have
\begin{align*}
\overline{\pi}_A(1)<\overline{\pi}_H(1)=\overline{\pi}_H(0)\le\overline{\pi}_A(0),
\end{align*}
where the equality holds by our assumption that $\alpha(\pi_H)=0$. Since $\overline{\pi}_A(1)\neq\overline{\pi}_A(0)$, we must have $\alpha(\pi_A)>0=\alpha(\pi_H)$, so by Theorem~\ref{thm:main}, $\overline{\pi}_A$ is not compliance-robustly fair, a contradiction.
\end{customproof}

\subsection{Proof of Results in Section \ref{section:performance}}
\label{appendix:proof_sec4}
To prove Theorem \ref{thm:existence}, we need the following lemmas. It follows by the construction of $\pi_B$ that it satisfies:
\begin{lemma}
\label{lem:pibkey}
We have $\pi_B(x,0)\ge\pi_H(x,0)$ and $\pi_B(x,1)\le\pi_H(x,1)$ for all $x\in\mathcal{X}$.
\end{lemma}

As we will see shortly, $\pi_B$ provides a constructive upper bound on the performance of any compliance-robustly fair policy, which will be useful for examining when the performance of $\pi_0$ exceeds that of $\pi_H$. We begin by noting that $\pi_B$ itself is compliance-robustly fair if it (in isolation) does not reduce fairness relative to $\pi_H$.
\begin{restatable}{lemma}{lemPib}
\label{lem:pib2}
If $\alpha(\pi_B)\le\alpha(\pi_H)$, then $\pi_B$ is compliance-robustly fair.
\end{restatable}
\begin{customproof}
By Lemma~\ref{lem:pibkey}, $\pi_B$ satisfies conditions (\ref{eqn:keycond2}) and (\ref{eqn:keycond3}) in Theorem~\ref{thm:main} by construction. If $\alpha(\pi_B)\le\alpha(\pi_H)$, then (\ref{eqn:keycond1}) also holds, so by Theorem~\ref{thm:main}, $\pi_B$ is compliance-robustly fair.
\end{customproof}
Furthermore, the next result shows that $\pi_B$ performs at least as well as the optimal compliance-robustly fair policy $\pi_0$.

\begin{restatable}{lemma}{lemPibTwo}
\label{lem:pib1}
We have $L(\pi_0)\ge L(\pi_B)$.
\end{restatable}
\begin{customproof}
It suffices to prove that $\pi_0$ has higher deviation than $\pi_B$, in which case the claim follows by Assumption~\ref{assump:loss}. We need to show that for all $x\in\mathcal{X}$ and $a\in\mathcal{A}$, we have
\begin{align}
\label{eqn:lem:pib1:cond}
\begin{cases}
\pi_0(x,a)\le\pi_B(x,a)&\text{if 
}\pi_B(x,a)\le\pi_*(x,a) \\
\pi_0(x,a)\ge\pi_B(x,a)&\text{if }\pi_B(x,a)\ge\pi_*(x,a).
\end{cases}
\end{align}
Now, consider a point $x\in u(0)$; in this case, we have
\begin{align*}
\pi_0(x,0)\ge\pi_H(x,0)\ge\pi_*(x,0),
\end{align*}
where the first inequality follows since $\pi_0$ is compliance-robustly fair so it satisfies (\ref{eqn:keycond2}), and the second since $x\in u(0)$. Since $\pi_B(x,0)=\pi_H(x,0)$ for $x\in u(0)$, (\ref{eqn:lem:pib1:cond}) holds. Next, consider a point $x\in\ell(1)$; in this case, we have
\begin{align*}
\pi_0(x,1)\le\pi_H(x,1)<\pi_*(x,1)
\end{align*}
where the first inequality follows since $\pi_0$ satisfies (\ref{eqn:keycond3}), and the second since $x\in\ell(1)$. Since $\pi_B(x,1)=\pi_H(x,1)$ for $x\in\ell(1)$, (\ref{eqn:lem:pib1:cond}) holds. Finally, if $x\not\in u(0)\cup\ell(1)$, then $\pi_B(x,a)=\pi_*(x,a)$ for all $a\in\mathcal{A}$, so (\ref{eqn:lem:pib1:cond}) holds. Thus, $\pi_0$ has higher deviation than $\pi_B$, so the claim follows.
\end{customproof}

Now, we prove Theorem \ref{thm:existence}.
\thmExistence*
\begin{customproof}
If $\alpha(\pi_B)\le\alpha(\pi_H)$, then by Lemma~\ref{lem:pib2}, $\pi_B$ is compliance-robustly fair; the assumptions in the theorem statement clearly imply that $L(\pi_B)<L(\pi_H)$, so the claim follows. Otherwise, we must have $\alpha(\pi_B)>\alpha(\pi_H)$. Furthermore, Lemma~\ref{lem:pibkey} implies that $\overline{\pi}_B(1)\le\overline{\pi}_H(1)$ and $\overline{\pi}_B(0)\ge\overline{\pi}_H(0)$. Together with Assumption~\ref{assump:group}, these three conditions imply that
\begin{align*}
\overline{\pi}_B(1)<\overline{\pi}_B(0).
\end{align*}
Intuitively, this might happen when the optimal policy satisfies $\overline{\pi}_*(1)<\overline{\pi}_*(0)$, but the human policy reverses this relationship.
To compensate, we can reduce the performance of $\overline{\pi}_B$ to ``shrink'' the gap between $\overline{\pi}_B(1)$ and $\overline{\pi}_B(0)$. In particular, consider scaling the decisions as follows:
\begin{align*}
  &\pi_{A,\lambda}(x,a)\\
&= \begin{cases}
\pi_H(x,a) & \text{if } x \in u(0) \cup \ell(1) \\
(1-\lambda)\pi_B(x,a) + \lambda\pi_H(x,a) & \text{otherwise}.
\end{cases}
\end{align*}

Note that $\pi_{A,0}=\pi_B$ and $\pi_{A,1}=\pi_H$. In addition, it is easy to see that $\pi_{A,\lambda}$ has strictly lower deviation than $\pi_H$ for all $\lambda\in[0,1)$ (strictness is due to Assumption~\ref{assump:probability} and our assumption on $\pi_H$ in the theorem statement). Next, by construction, for all $\lambda\in[0,1]$, we have $\pi_{A,\lambda}(x,1)\le\pi_H(x,1)$ and $\pi_{A,\lambda}(x,0)\ge\pi_H(x,0)$. Now, consider the function
\begin{align*}
g(\lambda)&=\overline{\pi}_{A,\lambda}(1)-\overline{\pi}_{A,\lambda}(0).
\end{align*}
By the above, we have
\begin{align*}
g(0)&=\overline{\pi}_B(1)-\overline{\pi}_B(0)\le0 \\
g(1)&=\overline{\pi}_H(1)-\overline{\pi}_H(0)\ge0.
\end{align*}
Thus, by the intermediate value theorem, there exists $\lambda^*\in[0,1]$ such that $g(\lambda^*)=0$. Since
\begin{align*}
g(1)=\overline{\pi}_H(1)-\overline{\pi}_H(0)=\alpha(\pi_H)\neq0,
\end{align*}
we know that $\lambda^*\neq1$, so $\lambda^*\in[0,1)$. Thus, $\pi_{A,\lambda^*}$ satisfies (\ref{eqn:keycond1}), (\ref{eqn:keycond2}), and (\ref{eqn:keycond3}), so by Theorem~\ref{thm:main}, it is compliance-robustly fair. In addition, since $\lambda_1^*\in[0,1)$, by the above, it has strictly lower deviation than $\pi_H$, so $L(\pi_{A,\lambda^*})<L(\pi_H)$. Thus, we have $L(\pi_0)\le L(\pi_{A,\lambda^*})<L(\pi_H)$, as claimed.
\end{customproof}

\subsection{Proof of Results in Section \ref{section:traditional_fairness}}
\label{appendix:proof_sec5}
\lemTraditional*
\begin{customproof}
We first show that existence of $\pi_A$ implies existence of $\pi$. By Theorem~\ref{thm:main}, $\pi_A$ satisfies
\begin{align*}
\overline{\pi}_A(1) &= \overline{\pi}_A(0) \\
\pi_H(x,0) &\leq \pi_A(x,0)
\qquad\qquad(\forall x\in\mathcal{X}) \\
\pi_A(x,1) &\leq \pi_H(x,1)
\qquad\qquad(\forall x\in\mathcal{X}).
\end{align*}
Now, let
\begin{align*}
\pi(x,a)=\begin{cases}
\max\{\pi_A(x,0),\pi_B(x,0)\}&\text{if }a=0 \\
\min\{\pi_A(x,1),\pi_B(x,1)\}
&\text{if }a=1.
\end{cases}
\end{align*}
By construction, $\pi$ satisfies (\ref{eqn:traditional:2}) and (\ref{eqn:traditional:3}). Furthermore, we have
\begin{align*}
\overline{\pi}(1)\le\overline{\pi}_A(1)=\overline{\pi}_A(0)\le\overline{\pi}(0),
\end{align*}
where the first inequality follows since $\pi(x,1)\le\pi_A(x,1)$ and the second since $\pi$ satisfies $\pi(x,0)\ge\pi_A(x,0)$. Thus, $\pi$ satisfies (\ref{eqn:traditional:1}). Finally, to show that $L(\pi)<L(\pi_H)$, it suffices to show that $\pi$ has lower or equal deviation compared to $\pi_A$, since this implies that $L(\pi)\le L(\pi_A)<L(\pi_H)$. To this end, recall that for all $x\in\mathcal{X}$ and $a\in\mathcal{A}$, we have $\pi_B(x,a)\in\{\pi_H(x,a),\pi_*(x,a)\}$. If $\pi_A(x,a)\neq\pi_H(x,a)$ and $\pi(x,a)\neq\pi_A(x,a)$, then we must $\pi(x,a)=\pi_B(x,a)$, so $\pi(x,a)\in\{\pi_H(x,a),\pi_*(x,a)\}$. In this case, we cannot have $\pi(x,a)=\pi_H(x,a)$, since either $a=0$ and $\pi(x,0)\ge\pi_A(x,0)>\pi_H(x,0)$, or $a=1$ and $\pi(x,1)\le\pi_A(x,1)<\pi_H(x,1)$. Thus, we must have $\pi(x,a)=\pi_*(x,a)$. In general, it follows that $\pi(x,a)\in\{\pi_A(x,a),\pi_*(x,a)\}$, which straightforwardly implies that $\pi$ has lower or equal deviation compared to $\pi_A$. The claim follows.

Next, we prove that existence of $\pi$ implies the existence of $\pi_A$. First, if $\overline{\pi}_B(1) \le \overline{\pi}_B(0)$, then the result follows from the proof of Theorem~\ref{thm:existence}, which shows that if $\overline{\pi}_B(1)\le\overline{\pi}_B(0)$, then there exists a compliance-robustly fair policy $\pi$ such that $\alpha(\pi)=0$. Thus, it suffices to consider the case $\overline{\pi}_B(1)>\overline{\pi}_B(0)$. In this case, by (\ref{eqn:traditional:2}) and (\ref{eqn:traditional:3}), we have
\begin{align*}
\pi(x,1) &\leq \min\{\pi_*(x,1), \pi_H(x,1)\} = \pi_B(x,1) \le \pi_*(x,1) \\
\pi(x,0) &\geq \max\{\pi_*(x,0), \pi_H(x,0)\} = \pi_B(x, 0) \ge \pi_*(x, 0).
\end{align*}
Thus, $\pi_B$ has lower or equal deviation compared to $\pi$,
so $L(\pi_B) \leq L(\pi) < L(\pi_H)$.
Consider
\begin{align*}
\pi_{A,\lambda}(x,a) = \lambda \pi(x,a) + (1-\lambda) \pi_B(x,a),
\end{align*}
where $\lambda \in [0,1]$. Note that $\pi_{A,0} = \pi_B$ and $\pi_{A,1} = \pi$. It is easy to see that $\pi_{A,\lambda}$ has lower or equal deviation compared to $\pi$, so $L(\pi_{A,\lambda}) \leq L(\pi) < L(\pi_H)$ for all $\lambda$.
Now, define
\begin{align*}
g(\lambda) = \overline{\pi}_{A,\lambda}(1) - \overline{\pi}_{A,\lambda}(0),
\end{align*}
so
\begin{align*}
g(0) &= \overline{\pi}_B(1) - \overline{\pi}_B(0) > 0\\
g(1) &= \overline{\pi}(1) - \overline{\pi}(0) < 0.
\end{align*}
By the intermediate value theorem, there exists $\lambda^* \in (0,1)$ such that $g(\lambda^*) = 0$. Then, we have $\alpha(\pi_{A,\lambda^*}) = 0$ and $L(\pi_{A,\lambda^*}) < L(\pi_H)$. It also directly follows from Theorem $\ref{thm:main}$ that $\pi_{A,\lambda^*}$ is compliance-robustly fair. Thus, $\pi_{A,\lambda^*}$ satisfies our desiderata, so the claim follows.
\end{customproof}

\propositionNoneExist*
\begin{customproof}
Let $\mathcal{X} = \{1\}$ be singleton; thus, we can omit it from our notation. Let
\begin{align*}
\mathbb{P}(a,y) = \begin{cases}
\frac{1}{2} (1-\epsilon) &\text{if }a=1\wedge y=1 \\
\frac{1}{2} \epsilon&\text{if }a=1\wedge y=0 \\
\frac{1}{2}\epsilon&\text{if }a=0\wedge y=1 \\
\frac{1}{2}(1-\epsilon)&\text{if }a=0\wedge y=0
\end{cases}
\end{align*}
for any $\epsilon\in(0,1/7]$. Let the loss be
\begin{align*}
L(\pi) &= \mathbb{E}[(\pi(a)-y)^2] \\
&= \frac{1}{2} [(1-\epsilon)(\pi(1)-1)^2 +\epsilon\pi(1)^2 \\
&\qquad +\epsilon(\pi(0)-1)^2+(1-\epsilon)\pi(0)^2 ].
\end{align*}
Then, it is easy to check that so that
\begin{align*}
\pi_*(a)=\begin{cases}
1-\epsilon&\text{if }a=1 \\
\epsilon&\text{if }a=0.
\end{cases}
\end{align*}
In addition, suppose that the human policy is
\begin{align*}
\pi_H(a)&=
\begin{cases}
1-\epsilon&\text{if }a=1 \\
\epsilon/2&\text{if }a=0.
\end{cases}
\end{align*}
In this case, $\pi_B=\pi_*$, and $\alpha(\pi_B)=\alpha(\pi_*)<\alpha(\pi_H)$, so by Theorem~\ref{thm:existence}, $\pi_B$ is compliance-robustly fair; in addition, it strictly improves performance, though it is itself unfair. Thus, $\Pi_{\text{fair}}\neq\varnothing$.

Next, we show that for any compliance-robustly fair policy $\pi$, if $\alpha(\pi)=0$, then $L(\pi)\ge L(\pi_H)$. Since $\mathcal{X}$ is singleton, we have $\overline{\pi}(a)=\pi(a)$, so $\alpha(\pi)=0$ implies $\pi(0)=\pi(1)$. Thus, it suffices to consider a policy $\pi(0)=\pi(1)=\beta$. For any such policy, the loss is
\begin{align*}
L(\pi) =&\frac{1}{2} \left[(1-\epsilon)(\beta-1)^2+\epsilon\beta^2+\epsilon(\beta-1)^2+(1-\epsilon)\beta^2\right] \\
=&\frac{1}{2} \left[(\beta-1)^2+\beta^2 \right],
\end{align*}
which is minimized when $\beta=1/2$, in which case $L(\pi)=1/4$. In contrast, we have

\begin{align*}
L(\pi_H)&= \frac{1}{2}[(1-\epsilon)(\epsilon)^2+\epsilon(1-\epsilon)^2\\
&\qquad +\epsilon(\epsilon/2 - 1)^2+(1-\epsilon)(\epsilon/2)^2 ] \\
&=\frac{1}{2}\left[(\epsilon/2)^2+2\epsilon(1-\epsilon)\right].
\end{align*}
It is easy to verify that when $\epsilon \in (0,\frac{1}{7}]$, we have $L(\pi_H) < \frac{1}{4} \leq L(\pi)$.


\end{customproof}

\subsection{Compliance Issues for General Fairness Conditions}
\label{general_fairness}

We define a general class of fairness criteria, subsuming demographic parity \cite{calders2009building,zliobaite2015relation} and equalized odds \cite{hardt2016equality,chen2023algorithmic}. We then show that, under this general class, fair policies are not necessarily compliance-robustly fair. Thus, in all cases, one must optimize separately for performance-improving compliance-robustly fair policies (as illustrated in Algorithm~\ref{fig:algorithm}).
%

We define a \emph{fairness criterion} as a function that takes a policy as input and outputs a value representing how fair the policy is. For example, $\alpha(\pi) = |\overline{\pi}(1) - \overline{\pi}(0)|$ quantifies fairness under the equality of opportunity criterion.

\begin{definition} 
\label{def:fairness} A \emph{fairness criterion} is a function $\varphi: \Pi \to \mathbb{R}_{\geq 0}$, where $\Pi$ is the space of all policies. Given $\varphi$, we say a policy $\pi\in\Pi$ is \emph{fairer} than another policy $\pi'\in\Pi$ if $\varphi(\pi) < \varphi(\pi')$. 
\end{definition}
Next, we extend our concept of a compliance-robustly fair policy to general fairness criteria.
\begin{definition} \label{def: general_compliance_robustness}
Given a human policy $\pi_H$ and an algorithmic policy $\pi_A$, we say that $\pi_A$ is \emph{compliance-robustly fair} with respect to $\pi_H$ if for every compliance function $c$, the resulting human-AI policy $\pi_C$ satisfies $\varphi(\pi_C) \leq \varphi(\pi_H)$.
\end{definition}

The following assumption characterizes the class of fairness criteria that are susceptible to selective compliance issues. That is, if a fairness criterion satisfies the assumption, there is tension between traditional fairness and compliance-robust fairness.

\begin{assumption}
\label{assump:fairnesscond}
Given a fairness condition $\varphi$, there exist policies $\pi_{\text{low}}$ and $\pi_{\text{high}}$, and a compliance function $c_0$, such that (i) we have
\begin{align*}
\varphi(\pi_{\text{low}}) < \varphi(\pi_{\text{high}}),
\end{align*}
(ii) the human-AI policy
\begin{align*}
\pi_C(x,a)=\begin{cases}
\pi_{\text{low}}(x,a)&\text{if }c_0(x,a)=1 \\
\pi_{\text{high}}(x,a)&\text{otherwise},
\end{cases}
\end{align*}
satisfies
\begin{align*}
\varphi(\pi_C) < \varphi(\pi_{\text{high}}),
\end{align*}
and (iii) the human-AI policy
\begin{align*}
\pi'_C(x,a)=\begin{cases}
\pi_{\text{high}}(x,a)&\text{if }c_0(x,a)=1 \\
\pi_{\text{low}}(x,a)&\text{otherwise}.
\end{cases}
\end{align*}
satisfies
\begin{align*}
\varphi(\pi'_C) < \varphi(\pi_{\text{high}}).
\end{align*}
\end{assumption}
In this assumption, condition (i) says that according to $\varphi$, the policies $\pi_{\text{low}}$ and $\pi_{\text{high}}$ are increasingly unfair. Then, condition (ii) says that if $\pi_{\text{high}}$ is the human policy and $\pi_{\text{low}}$ is the AI policy, then the resulting human-AI policy under the compliance function $c_0$ is strictly fairer than the human policy $\pi_{\text{high}}$. Finally, condition (iii) says that if $\pi_{\text{low}}$ is the human policy and $\pi_{\text{high}}$ is the AI policy, then the resulting human-AI policy under $c_0$ is again strictly more fair than $\pi_{\text{high}}$. In fact, condition (i) is not necessary, but we include it since it adds intuition---the human-AI policy can be thought of as moving closer to $\pi_{\text{low}}$ from $\pi_{\text{high}}$ in both cases.

Intuitively, conditions (ii) and (iii) say that there exist two policies $\pi_{\text{low}}$ and $\pi_{\text{high}}$ with different fairness levels such that either of the two human-AI policies formed by combining them has fairness strictly less than $\pi_{\text{high}}$. These conditions are met by a wide range of algorithmic fairness definitions; later in this section, we will show that two widely-used fairness definitions---demographic parity and equalized odds---satisfy it.

Next, we show that any fairness definition satisfying Assumption~\ref{assump:fairnesscond} is vulnerable to the selective compliance problem. This result demonstrates the pervasive nature of the selective compliance problem; as a result, there exists an inherent tension between traditional fairness and compliance-robust fairness for a broad class of fairness definitions.

\begin{theorem}
\label{thm:impossibility}
For any fairness condition $\varphi$ satisfying Assumption~\ref{assump:fairnesscond}, there exists a human policy $\pi_H$ and an algorithmic policy $\pi_A$ such that $\varphi(\pi_A)\le\varphi(\pi_H)$ but $\pi_A$ is not compliance-robustly fair for $\pi_H$.
\end{theorem}
\begin{customproof}
We show that it is always possible to construct a human-AI policy $\pi_C$ that is less fair than the human-alone policy $\pi_H$ under Assumption~\ref{assump:fairnesscond}, even though the AI policy $\pi_A$ is fairer than the human-alone policy $\pi_H$.

Let $\pi_{\text{low}}$, $\pi_{\text{high}}$, and $c_0$ be as defined in Assumption~\ref{assump:fairnesscond}, and consider the policy
\begin{align*}
\pi_1(x,a) =
\begin{cases}
\pi_{\text{high}}(x,a) &\text{if } c_0(x,a) = 1 \\
\pi_{\text{low}}(x,a) &\text{otherwise}.
\end{cases}
\end{align*}
By Assumption~\ref{assump:fairnesscond}, $\varphi(\pi_1) < \varphi(\pi_{\text{high}})$. Also, consider the policy
\begin{align*}
\pi_2(x,a)=\begin{cases}
\pi_\text{low}(x,a)&\text{if }c_0(x,a)=1 \\
\pi_{\text{high}}(x,a)&\text{otherwise}.
\end{cases}
\end{align*}
By Assumption~\ref{assump:fairnesscond}, $\varphi(\pi_2)<\varphi(\pi_{\text{high}})$. Now, if $\varphi(\pi_1)\le\varphi(\pi_2)$, then consider
\begin{align*}
\pi_C^{(1)}(x,a)=\begin{cases}
\pi_1(x,a)&\text{if }c_0(x,a)=1 \\
\pi_2(x,a)&\text{otherwise}.
\end{cases}
\end{align*}
Note that $\pi_C^{(1)}=\pi_{\text{high}}$ since $\pi_C^{(1)}(x,a)=\pi_1(x,a)=\pi_{\text{high}}(x,a)$ if $c_0(x,a)=1$ and $\pi_C^{(1)}(x,a)=\pi_2(x,a)=\pi_{\text{high}}(x,a)$ otherwise. Thus, $\varphi(\pi_2)<\varphi(\pi_{\text{high}})=\varphi(\pi_C^{(1)})$. Taking $\pi_A = \pi_1$ and $\pi_H = \pi_2$, we have $\varphi(\pi_1) \leq \varphi(\pi_2)$, but $\pi_1$ is not compliance-robustly fair for $\pi_2$ because $\varphi(\pi_C^{(1)}) > \varphi(\pi_2)$.

Otherwise, we have $\varphi(\pi_1)>\varphi(\pi_2)$. Let
\begin{align*}
\pi_C^{(2)}(x,a)=\begin{cases}
\pi_2(x,a)&\text{if }\tilde{c}_0(x,a)=1 \\
\pi_1(x,a)&\text{otherwise}.
\end{cases}
\end{align*}
where
\begin{align*}
\tilde{c}_0(x,a)=1-c_0(x,a).
\end{align*}
Similar to before, we have $\pi_C^{(2)}=\pi_{\text{high}}$. Thus, $\varphi(\pi_1)<\varphi(\pi_{\text{high}})=\varphi(\pi_C^{(2)})$. Taking $\pi_A = \pi_2$ and $\pi_H = \pi_1$, we have $\varphi(\pi_2) < \varphi(\pi_1)$, but $\pi_2$ is not compliance-robustly fair for $\pi_1$ because $\varphi(\pi_C^{(2)}) > \varphi(\pi_1)$.
\end{customproof}

\paragraph{Demographic parity} Now, we show that demographic parity satisfies Assumption~\ref{assump:fairnesscond}, implying that it suffers from compliance-related problems. In particular, redefine the following:
\begin{align*}
\overline{\pi}(a) = \sum_{x\in\mathcal{X}} \pi(x,a)\mathbb{P}(x\mid a),
\end{align*}
so demographic parity is given by $\alpha_D(\pi)=|\bar{\pi}(1)-\bar{\pi}(0)|$.
We need to establish a setting for which the two policies and the compliance function in Assumption~\ref{assump:fairnesscond} exist. Let $\mathcal{X} = \{1\}$ be singleton; then, we can omit it from our notation. Next, we construct $\pi_{\text{high}}$ and $\pi_{\text{low}}$ as follows:
\begin{align*}
\pi_{\text{low}}(a) &= \begin{cases}
\frac{1}{2} + \epsilon \quad \text{if $a = 1$} \\
\frac{1}{2} - \epsilon \quad \text{if $a = 0$}
\end{cases}\\
\pi_{\text{high}}(a) &= \begin{cases}
\frac{1}{2} + 3 \epsilon \quad \text{if $a = 1$} \\
\frac{1}{2} - 2 \epsilon \quad \text{if $a = 0$},
\end{cases}
\end{align*}
where $\epsilon \in (1/6,1/4)$. Also, consider the compliance function:
\begin{align*}
c_0(a) = 
\begin{cases}
1 \quad \text{if $a = 1$} \\
0 \quad \text{if $a = 0$},
\end{cases}
\end{align*}
which implies $\pi_C$ and $\pi'_C$ are as follows:
\begin{align*}
\pi_C(a) &= 
\begin{cases}
\frac{1}{2} + \epsilon & \begin{aligned} \text{if } a &= 1 \end{aligned} \\
\frac{1}{2} - 2\epsilon & \begin{aligned} \text{if } a &= 0 \end{aligned}
\end{cases} \\
\pi'_C(a) &= 
\begin{cases}
\frac{1}{2} + 3\epsilon & \begin{aligned} \text{if } a &= 1 \end{aligned} \\
\frac{1}{2} - \epsilon & \begin{aligned} \text{if } a &= 0 \end{aligned}
\end{cases}
\end{align*}
With these definitions, it is easy to see that Assumption~\ref{assump:fairnesscond} is satisfied.

\paragraph{Equalized Odds}

The case of equalized odds is similar to that of equal opportunities. Redefine the following:
\begin{align*}
\overline{\pi}(a, y) = \sum_{x\in\mathcal{X}}\pi(x,a)\mathbb{P}(x|a, y).
\end{align*}
Then, equalized odds can be defined as follows:
\begin{align*}
\varphi(\pi) = \sup_{y \in \{0,1\}}|\overline{\pi}(1,y) - \overline{\pi}(0,y)|.
\end{align*}
As before, consider $\mathcal{X} = \{1\}$ be singleton, then we can omit it from our notation. Note that $\overline{\pi}(1,y) = \pi(1)$ and $\overline{\pi}(0,y) = \pi(0)$.  Next, we define $\pi_{\text{high}}$ and $\pi_{\text{low}}$ as follows:
\begin{align*}
\pi_{\text{low}}(a) &= \begin{cases}
\frac{1}{2} + \epsilon \quad \text{if $a = 1$} \\
\frac{1}{2} - \epsilon \quad \text{if $a = 0$}
\end{cases}\\
\pi_{\text{high}}(a) &= \begin{cases}
\frac{1}{2} + 3 \epsilon \quad \text{if $a = 1$} \\
\frac{1}{2} - 2 \epsilon \quad \text{if $a = 0$},
\end{cases}
\end{align*}
where $\epsilon \in (1/6,1/4)$. Also, consider the compliance function:
\begin{align*}
c_0(a) = 
\begin{cases}
1 \quad \text{if $a = 1$} \\
0 \quad \text{if $a = 0$},
\end{cases}
\end{align*}
which implies that $\pi_C$ and $\pi'_C$ are as follows:
\begin{align*}
\pi_C(a) &= 
\begin{cases}
\frac{1}{2} + \epsilon & \begin{aligned} \text{if } a &= 1 \end{aligned} \\
\frac{1}{2} - 2\epsilon & \begin{aligned} \text{if } a &= 0 \end{aligned}
\end{cases}\\
\pi'_C(a) &= 
\begin{cases}
\frac{1}{2} + 3\epsilon & \begin{aligned} \text{if } a &= 1 \end{aligned} \\
\frac{1}{2} - \epsilon & \begin{aligned} \text{if } a &= 0 \end{aligned}
\end{cases}
\end{align*}
Again, it is easy to see that Assumption \ref{assump:fairnesscond} is satisfied.

\section{Experimental Details} \label{exp-details}

\paragraph{Sample Selection.} Following the setup of \citet{stevenson2022algorithmic}, we restrict the sample to defendants that are eligible for the non-violent risk assessment tool, which is our population of interest---we select defendants that (i) committed a drug, larceny, or fraud offense, (ii) do not have a history of violent offenses, and (iii) are considered for a prison or jail sentence. Then, we augment the criminal sentence records by merging it with defendants' demographic information obtained from the Virginia Court Data website. We also restrict to Non-Hispanic White and Black defendants. Since we use data from before the introduction of the risk assessment tool to learn $\pi_H$, judges who appear only after the tool's implementation are excluded from our analyses.

\paragraph{Defendant Covariates.} We use ``Defendant Sex'', ``Defendant Age'', ``Defendant Race'', ``Defendant in Youthful Offender Program'', ``Charge Type'', ``Mandatory Minimum Sentence'', ``Recommend Prison'', ``First Offender'', ``Recommended Sentence Length'', and ``Primary Offenses''.

\paragraph{Estimating the judges' policies.} We leverage guidelines-recommended sentences to infer judges' perceived recidivism risk. The guidelines provide judges with a range of suitable sentences (e.g., 6 months to 2 years), the midpoint of which is defined as the ``guidelines-recommended sentence.'' Following \citet{stevenson2022algorithmic}, we consider the judge to perceive an offender to have a low recidivism risk if (i) the guideline-recommended sentence is prison (more than 12 months), but the judge assigns a sentence of 6 months or less of jail time, or (ii) the guideline-recommended sentence is jail (less or equal to 12 months), but the judge assigns a sentence of zero (i.e., not incarcerated at all).

\paragraph{Human-AI Collaborative Policy Construction.} We first estimate the optimal performance-maximizing policy $\pi_A^*$, our compliance-robustly fair policy $\pi_A^{\text{robust}}$, and the performance-maximizing traditionally fair (i.e., satisfying Equality of Opportunity) policy $\pi_A^{\text{trad-fair}}$ for each judge (based on their estimated $\pi_H$) using data prior to the deployment of the risk assessment tool. 

To learn $\pi_A^*$ and $\pi_A^{\text{trad-fair}}$, we require the true outcome $y$ for defendants. Consistent with VCSC's definitions, we label any defendant that receives another felony conviction within a three-year window after their release as a recidivist. We then train a gradient boosted decision tree~\citep{ke2017lightgbm} to predict whether a defendant is a recidivist based on the same observed defendant covariates, yielding $\pi_A^*$. For $\pi_A^{\text{trad-fair}}$, we use the methods proposed by~\citet{weerts2023fairlearn} to enforce the Equality of Opportunity fairness constraint. Table~\ref{table:data_sources} summarizes the data sources used to train each policy.

\begin{table}[t]
\centering
\begin{tabular}{lcc}
\hline
\textbf{Policy/Function} & \textbf{Before-deployment Data} & \textbf{Post-deployment Data} \\
\hline
$\pi_H$ & $\checkmark$ &  \\
$\pi_A^{\text{actual}}$ &  & $\checkmark$ \\
$\pi_A^{*}$ & $\checkmark$ &  \\
$\pi_A^{\text{trad-fair}}$ & $\checkmark$ &  \\
$\pi_A^{\text{robust}}$ & $\checkmark$ &  \\
Compliance functions &  & $\checkmark$ \\
\hline
\end{tabular}
\caption{\textbf{Data used for policy estimation.} 
$\pi_A^{\text{actual}}$ and the compliance functions are estimated using post-deployment data (i.e., data after 2002). 
All other policies are learned from data collected before the tool's deployment.}
\label{table:data_sources}
\end{table}

Then, we have the following four policies:
\begin{align*}
\pi^{\text{actual}}_C(x,a)
&=\begin{cases}
\pi_A^{\text{actual}}(x,a)&\text{if }c(x,a)=1 \\
\pi_H(x,a)&\text{otherwise}.
\end{cases}
\end{align*}
\begin{align*}
\pi^{\text{robust}}_C(x,a)
&=\begin{cases}
\pi^{\text{robust}}_A(x,a)&\text{if }c(x,a)=1 \\
\pi_H(x,a)&\text{otherwise}.
\end{cases}
\end{align*}
\begin{align*}
\pi^{*}_C(x,a)
&=\begin{cases}
\pi_A^{*}(x,a)&\text{if }c(x,a)=1 \\
\pi_H(x,a)&\text{otherwise}.
\end{cases}
\end{align*}
\begin{align*}
\pi^{\text{trad-fair}}_C(x,a)
&=\begin{cases}
\pi_A^{\text{trad-fair}}(x,a)&\text{if }c(x,a)=1 \\
\pi_H(x,a)&\text{otherwise},
\end{cases}
\end{align*}

\paragraph{Metrics.} To evaluate performance, we compute the average loss as follows:
\begin{align*}
L(\pi_H) - L(\pi_C) &= \frac{1}{N}\sum_{i=1}^N \ell(\pi_H(x_i,a_i),y_i) \\
&\qquad -\frac{1}{N}\sum_{i=1}^N \ell(\pi_C(x_i,a_i),y_i) ,
\end{align*}
where $N$ is the number of samples in our evaluation dataset. The outcome $y_i$ indicates whether the defendant $i$ in fact recidivates (which we observe in the data). Note that a positive difference in average loss indicates an improvement in performance over the judges' policy.  Similarly, we evaluate fairness using the following metric:
\begin{align*}
\alpha(\pi_H) - \alpha(\pi_C),
\end{align*}
where $\alpha(\pi)$ is the slack in group fairness for $\pi$. In this case, a positive difference indicates that $\pi_C$ improves equity over the judges' policy.

\begin{figure}[t]
\centering
\begin{subfigure}{0.43\textwidth}
\centering
\includegraphics[width=\linewidth]{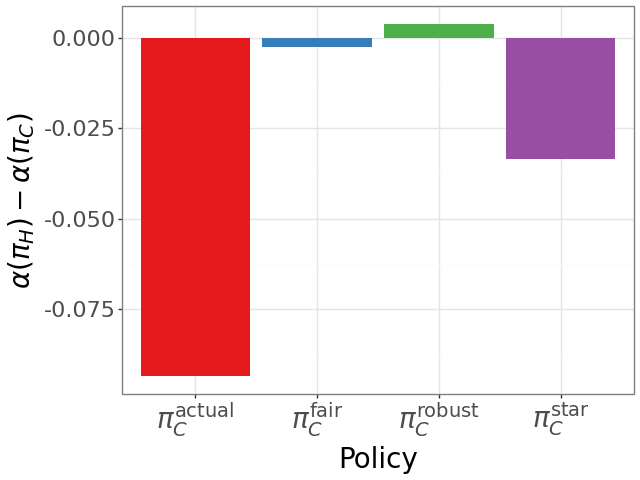} 
\caption{Fairness of four human-AI policies under the observed problematic compliance function}
\label{subfig:fairness_worst_case}
\label{fig:worst_case_compliance_agg}
\end{subfigure}
\hfill
\begin{subfigure}{0.43\textwidth}
\centering
\includegraphics[width=\linewidth]{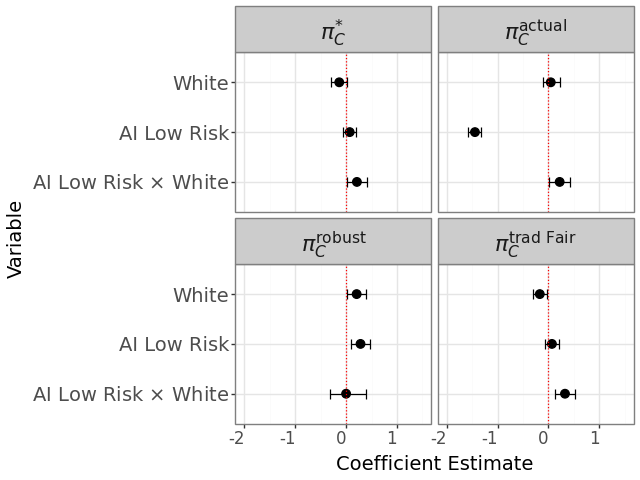} 
\caption{Regression coefficients from regressing judges' compliance decisions on defendants' race and AI's recommendation}
\label{subfig:compliance_reg}
\label{fig:worst_case_compliance_reg}
\end{subfigure}
\caption{\scriptsize In the left panel, we show the fairness comparison of $\pi_C^{\text{actual}}$, $\pi_C^{\text{robust}}$, $\pi_C^*$ and $\pi_C^{\text{trad-fair}}$ for a problematic compliance function. In the right panel, we present the regression coefficients from the regression specification in Section~\ref{exp-results}. The error bars are 95\% bootstrapped confidence intervals.}
\label{}
\end{figure}


\begin{figure}[h!]
\centering
\begin{subfigure}{0.43\textwidth}
\centering
\includegraphics[width=\linewidth]{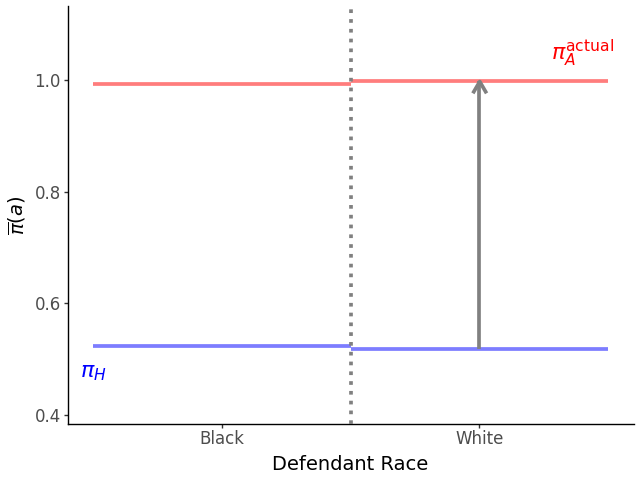} 
\caption{$\pi_A^{\text{actual}}$}
\end{subfigure}
\hfill
\begin{subfigure}{0.43\textwidth}
\centering
\includegraphics[width=\linewidth]{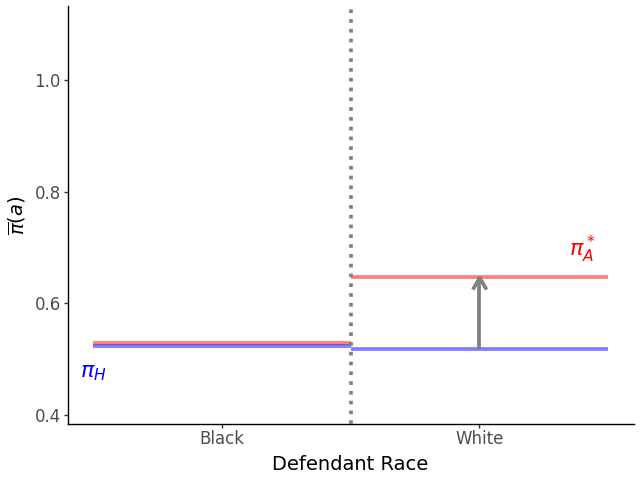} 
\caption{$\pi_A^{*}$}
\end{subfigure}
\hfill
\begin{subfigure}{0.43\textwidth}
\centering
\includegraphics[width=\linewidth]{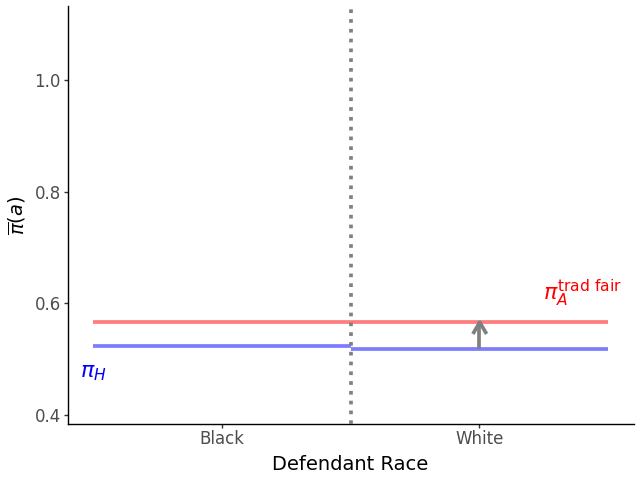} 
\caption{$\pi_A^{\text{trad-fair}}$}
\end{subfigure}
\hfill
\begin{subfigure}{0.43\textwidth}
\centering
\includegraphics[width=\linewidth]{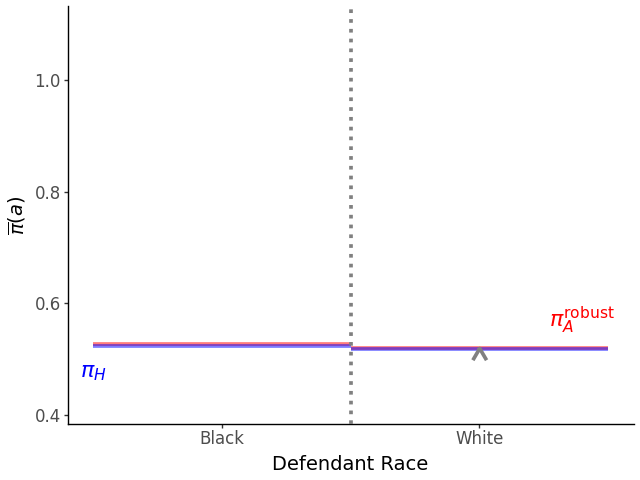} 
\caption{$\pi_A^{\text{robust}}$}
\end{subfigure}
\caption{\scriptsize Policy depictions for a defendant subgroup that evokes increased unfairness: [female, 40-50 years old, guideline recommends prison, non-first offender, drug-related offenses]. $\pi_A^{\text{robust}}$ preserves fairness by imitating the human policy.}
\label{fig:mech2}
\end{figure}

\paragraph{Problematic Compliance Patterns.} We focus on judges who exhibit fairness deterioration, as shown in Figure~\ref{fig:judge_level_analyses}---i.e., $\alpha(\pi_H) - \alpha(\pi_C) < 0$ across $\pi_A^{*}$, $\pi_A^{\text{actual}}$, and $\pi_A^{\text{trad-fair}}$, yielding 21 judges. From these judges, we derive a single ``problematic compliance function'', $c_{\text{problem}}$, by averaging individual judges' compliance functions. Similarly, we compute the ``average human-alone policy,'' $\pi^{\text{Ave}}_H$, by averaging their individual human-alone policies.

Using the problematic compliance function $c_{\text{problem}}$ and the average human-alone policy $\pi^{\text{Ave}}_H$, we simulate the four human-AI policies: $\pi_C^{*}$, $\pi_C^{\text{robust}}$, $\pi_C^{\text{actual}}$, and $\pi_C^{\text{trad-fair}}$. We compare their fairness and present the results in Figure~\ref{subfig:fairness_worst_case}. The vertical axis represents the unfairness level, $\alpha(\pi_H) - \alpha(\pi_C)$. A negative value indicates that the human-AI policy $\pi_C$ is less fair than the human-alone policy $\pi_H$. Indeed, all human-AI policies, except the compliance-robustly fair policy, reduce fairness.

In the regression presented in Section~\ref{exp-results}, the parameter $\beta_3$ captures our quantity of interest---a positive value indicates that judges comply more often for White defendants when the algorithmic recommendation is more lenient than their independent decisions, suggesting that racial disparities are exacerbated under algorithmic advice. We run this regression for each of the four human-AI policies. As shown in Figure~\ref{subfig:compliance_reg}, the estimated $\beta_3$ is positive and statistically significant for all human-AI policies ($\pi_C^{\text{actual}}$, $\pi_C^{\text{trad-fair}}$, and $\pi_C^{*}$), indicating that judges' compliance behaviors exacerbate existing racial biases under these policies. In contrast, our compliance-robustly fair policy ($\pi_A^{\text{robust}}$) effectively guards against such problematic compliance behaviors.

As discussed in Section~\ref{section:performance}, the algorithmic policy cannot further advantage the advantaged group (in this case, Whites) than the human-alone policy without risking increased disparities for problematic compliance patterns. We identify a defendant subgroup that experiences the most significant fairness deterioration under our ``problematic compliance function''---specifically, 40-50 year old females who are not first offenders, are charged with drug-related offenses, and are recommended prison time based on VCSC guidelines. In Figure~\ref{fig:mech2}, we illustrate the compliance implications for different algorithmic advice strategies. Our compliance-robust policy is the only one that preserves overall fairness by imitating the human policy for the advantaged subgroup, as in the construction of $\pi_B$.

\end{APPENDICES}

\end{document}